\documentclass[10pt,twocolumn,letterpaper]{article}

\PassOptionsToPackage{numbers, sort&compress}{natbib}
\usepackage[pagenumbers]{cvpr}      %

\usepackage[utf8]{inputenc} %
\usepackage[T1]{fontenc}    %
\usepackage{url}            %
\usepackage{booktabs}       %
\usepackage{amsfonts}       %
\usepackage{nicefrac}       %
\usepackage{microtype}      %

\usepackage[table,x11names,dvipsnames]{xcolor}
\usepackage{amsmath}
\usepackage{amssymb}
\usepackage{mathtools}
\usepackage{amsthm}

\usepackage{amsmath,amsfonts,bm}

\def\1{\bm{1}}

\def\sp{space}

\def\rvc{{\mathbf{c}}}

\def\rvx{{\mathbf{x}}}

\def\va{{\bm{a}}}
\def\vb{{\bm{b}}}
\def\vc{{\bm{c}}}

\def\vg{{\bm{g}}}

\def\vn{{\bm{n}}}

\def\vx{{\bm{x}}}

\def\mI{{\bm{I}}}

\DeclareMathAlphabet{\mathsfit}{\encodingdefault}{\sfdefault}{m}{sl}
\SetMathAlphabet{\mathsfit}{bold}{\encodingdefault}{\sfdefault}{bx}{n}
\newcommand{\tens}[1]{\bm{\mathsfit{#1}}}

\def\tG{{\tens{G}}}

\def\gG{{\mathcal{G}}}

\def\gL{{\mathcal{L}}}

\def\gN{{\mathcal{N}}}

\def\sR{{\mathbb{R}}}

\DeclareMathOperator*{\argmin}{arg\,min}

\newcommand*{\ShowNotes}{} %
\definecolor{darkred}{rgb}{0.7,0.1,0.1}
\definecolor{darkgreen}{rgb}{0.1,0.7,0.1}
\definecolor{cyan}{rgb}{0.7,0.0,0.7}
\definecolor{dblue}{rgb}{0.2,0.2,0.8}
\definecolor{maroon}{rgb}{0.76,.13,.28}
\definecolor{burntorange}{rgb}{0.81,.33,0}
\definecolor{tealblue}{rgb}{0.212,0.459, 0.533}

\definecolor{mypink}{rgb}{0.93359375, 0.62109375, 0.83984375}

\definecolor{pp}{rgb}{0.43921569, 0.18823529, 0.62745098}
\definecolor{rr}{rgb}{0.5254902 , 0.00784314, 0.12941176}
\definecolor{bb}{rgb}{0.09019608, 0.23529412, 0.37647059}
\definecolor{yy}{rgb}{0.49803922, 0.3372549 , 0.0}
\definecolor{gg}{rgb}{0.02352941, 0.3372549 , 0.17647059}

\ifdefined\ShowNotes
  \newcommand{\colornote}[3]{{\color{#1}\bf{#2: #3}\normalfont}}
\else
  \newcommand{\colornote}[3]{}
\fi

\usepackage{symbols}

\definecolor{mybrown}{rgb}{0.87058824, 0.56078431, 0.01960784}
\definecolor{myblue}{rgb}{0.3372549 , 0.70588235, 0.91372549}
\definecolor{mypurple}{rgb}{0.8, 0.47058824, 0.7372549 }
\definecolor{myorange}{rgb}{0.835, 0.368, 0}
\definecolor{mygreen}{rgb}{0.00784314, 0.61960784, 0.45098039}
\definecolor{mygt}{rgb}{0.0078125 , 0.57421875, 0.40625}
\definecolor{mysp}{rgb}{0.84765625, 0.515625  , 0.0234375}
\definecolor{mycitecolor}{rgb}{0,0.08,0.45}
\definecolor{mygr}{rgb}{0.9607,0.9607,0.9607}
\definecolor{myoo}{rgb}{0.992,0.9176,0.9019}

\definecolor{myrr}{HTML}{AE031A}
\definecolor{mybb}{HTML}{0155B3}

\definecolor{cvprblue}{rgb}{0.21,0.49,0.74}
\usepackage[pagebackref,breaklinks,colorlinks,citecolor=cvprblue]{hyperref}

\newcommand{\myparagraph}[1]{\vspace*{2pt}{\bf\noindent #1}}

\def\confName{CVPR}
\def\confYear{2024}

\title{AmbiGen: Generating Ambigrams from Pre-trained Diffusion Model}

\author{
Boheng Zhao\textsuperscript{1}\quad\quad Rana Hanocka\textsuperscript{2}\quad\quad  Raymond A. Yeh\textsuperscript{1}\\
\textsuperscript{1}Dept. of CS, Purdue University\quad\quad
\textsuperscript{2}Dept. of CS, University of Chicago\\
}

\begin{document}

\twocolumn[{%
\renewcommand\twocolumn[1][]{#1}%
\maketitle
\vspace{-0.8cm}
\begin{center}
\includegraphics[width=\linewidth]{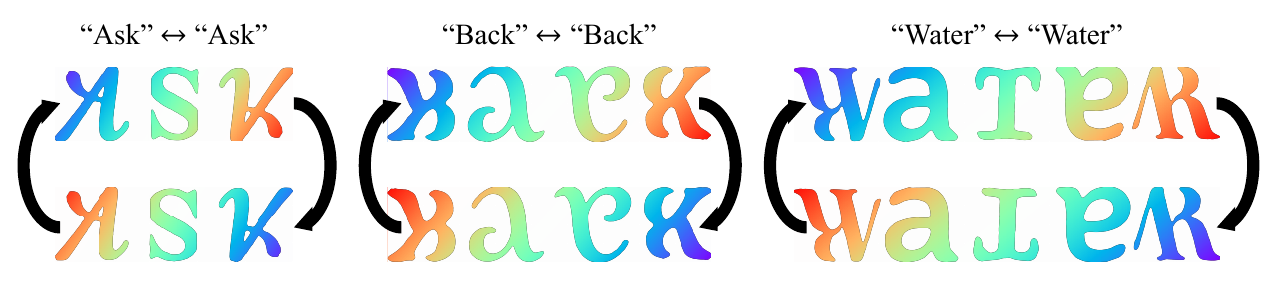}
\end{center}
\vspace{-0.65cm}
\captionof{figure}{Ambigrams designed by our proposed method. Observe that the words can be read the same after a rotation of 180 degrees. The words are shown with color gradients to better visualize the correspondence before and after the rotation.
}
\label{fig:teaser}
\vspace{0.7cm}
}]

\begin{abstract}
    Ambigrams are calligraphic designs that have different meanings depending on the viewing orientation. Creating ambigrams is a challenging task even for skilled artists, as it requires maintaining the meaning under two different viewpoints at the same time. In this work, we propose to generate ambigrams by distilling a large-scale vision and language diffusion model, namely DeepFloyd IF, to 
    optimize the letters' outline for legibility in the two viewing orientations. Empirically, we demonstrate that our approach outperforms existing ambigram generation methods. On the 500 most common words in English, our method achieves more than an 11.6\% increase in word accuracy and at least a 41.9\% reduction in edit distance. 
\end{abstract}
\section{Introduction}

Through meticulously designed fonts, ambigrams are words that can be viewed from different orientations. The most common are ``rotational ambigrams" which can be viewed after a 180-degree rotation, \eg, the word ``SWIMS" naturally reads the same when viewed both right-side up and upside down. 

Designing ambigrams is challenging and time-consuming as it requires the ``designer to solve a visual puzzle''~\cite{adobeambi}. While there are tutorials on how to design ambigrams~\cite{adobeambi,ambiwiki}, the instructions only contain general guidelines and many tedious details need to be implemented by a designer. Ultimately, making high-quality and effective ambigrams depends on a designer's understanding of calligraphy, symmetry patterns, and how to trade off the legibility of words for different orientations. 

Existing generators~\cite{makeambigram,ambigramania} construct ambigrams using a letter-to-letter approach. That is, each glyph in a word needs to look like two letters from different orientations. For rotational ambigrams, the font contains $26\times 26= 676$ glyphs which map between all pairs of %
letters %
in the alphabet. Conventionally, these ambigram fonts are designed by artists~\cite{makeambigram,ambigramania}. More recently, AmbiDream~\cite{ambidream} and AmbiFusion~\cite{shirakawa2023ambigram} leverage deep neural networks to aid the generation of ambigrams in the pixel space. 
However, they are limited to only designing ambigrams at a letter level and did not benchmark the performance at the word level. %

In this work, we propose a method to generate ambigrams by leveraging recent developments in text-to-image foundation models. We formulate the design process of an ambigram as an optimization problem where we directly optimize the control points of the Bezier curves representing an ambigram. At a high level, 
to maintain the generation's legibility,
the objective function is based on DeepFloyd IF~\cite{deep-floyd-if}, which employs a 
T5-XXL~\cite{t5xxl} text encoder to better capture information from the input text prompt.
We also incorporate a style loss to control the font styles. In~\figref{fig:teaser}, we show the results of rotational ambigrams designed by our method. 

\begin{figure*}[t]
\centering
\vspace{-0.07cm}
\includegraphics[width=0.97\linewidth]{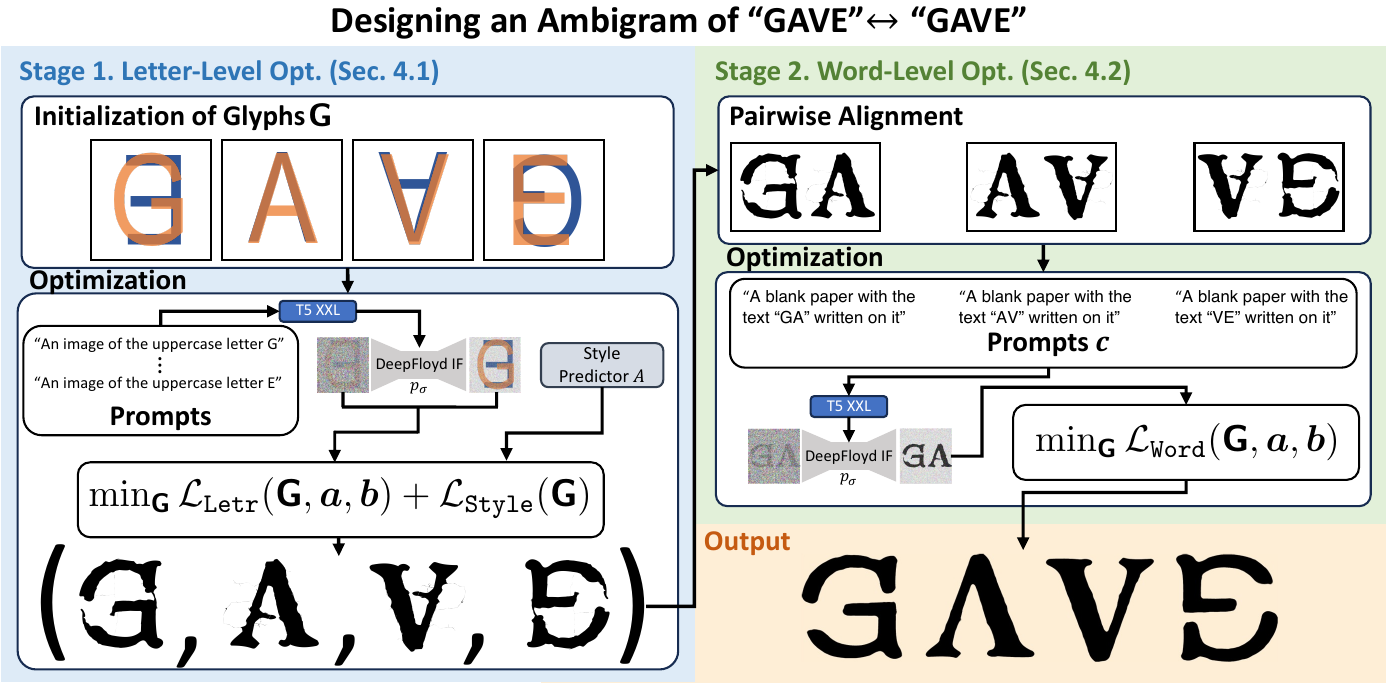}
\vspace{-0.2cm}
\caption{Illustration of the overall approach for designing an ambigram of ``GAVE'' $\leftrightarrow$ ``GAVE''. Our approach first performs letter-level optimization to design the individual glyphs (See~\secref{sec:letter}). Next, we optimize the legibility between pairs of glyphs from the first stage (See~\secref{sec:word}) leading to the final ambigram output. Note: the colors in the initialization are only for illustration purposes.
}
\vspace{-0.35cm}
\label{fig:pipeline}
\end{figure*}
To validate our approach, we use Optical Character Recognition (OCR)~\cite{trocr} to evaluate the legibility of the generated ambigrams under the two viewing (rotated by 180 degrees) orientations. %
Furthermore, we provide qualitative results and conduct a user study. Overall, we observe that our proposed method convincingly outperforms, both quantitative and qualitatively, the existing ambigram generation baselines.

{\bf \noindent Our contributions are as follows:}
\begin{itemize}
\item We propose an optimization framework, based on a pre-trained diffusion model, for ambigram generation.
\item To the best of our knowledge, we are the first to 
benchmark word-level ambigrams and generate them via deep neural networks.
\item Our method achieves more than 11.2\% absolute increase in word accuracy on the generated ambigrams compared to the existing ambigram generations, including ones designed by artists.
\end{itemize}

\section{Related work}\label{sec:rel}
{\noindent\bf Diffusion models and applications.}
Vision and language diffusion models, \eg, Stable Diffusion~\cite{rombach2022high}, DALL-E 2~\cite{ramesh2022hierarchical}, Imagen~\cite{saharia2022photorealistic}, have achieved impressive capabilities in language conditioned image generation. With these advancements, many applications have arisen by using a trained diffusion model as a prior term,~\eg, generation of 3D models ~\cite{dreamfusion,wang2020attribute2font}, and vector art~\cite{jain2023vectorfusion}, image restoration~\cite{kawar2022denoising,xie2023smartbrush,lugmayr2022repaint}, shape reconstruction~\cite{liu2023zero}, \etc. This work investigates how to use a pre-trained diffusion model for generating ambigrams.

\myparagraph{Font generation with AI.}
Many works have studied how to generate fonts using generative AI, commonly formulated as a probabilistic generative model in pixel space~\cite{azadi2018multi,hayashi2019glyphgan,gao2019artistic,wang2020attribute2font} or the space of vector fonts~\cite{reddy2021multi,liu2022learning,lopes2019learned,carlier2020deepsvg,wang2021deepvecfont,liu2023dualvector,cao2023svgformer,wang2023deepvecfont,chen2023joint}. Beyond fonts synthesis,~\citet{wang2022aesthetic} studied how to generate text logos focusing on glyph placements. More closely related to this work, 
WordAsImage by~\citet{Word-As-Image} uses score distillation of StableDiffusion on vector fonts to design semantic typography. While this work also distills a pre-trained diffusion model, we focus on ambigram generation. For this, we introduce several novel components tailored for ambigram generation, including how to initialize the optimization, which pre-trained diffusion model to use, and how to perform word-level optimization.

\myparagraph{Ambigram generation with AI.}
Ambidream~\cite{ambidream} proposed to generate ambigrams by distilling a pre-trained letter classifier to update the pixel values. Differently, we distill a large-scale diffusion model to update the outline of each glyph. Very recently, Ambifusion~\cite{shirakawa2023ambigram} proposed to train a diffusion model for ambigram generation by preparing a dataset of images containing individual alphabets manually cleaned from the MyFonts dataset~\cite{chen2019large}. In contrast, our method does not require the training of a text-specific diffusion model. Finally, both Ambidream and Ambifusion are limited to generating single-letter ambigrams and do not consider word-level interaction in their method or evaluation.

\section{Preliminaries}
We review the necessary concepts to understand our approach and to establish the notation.

\myparagraph{Glyph representation.} 
Glyphs are commonly stored in vector form, \eg, via Bézier curves that capture the outline, as the representation can be scaled to an arbitrary resolution without losing details. In more detail, a glyph $\gG = \{\vg_i\}$ is denoted as a set of control points $\vg_i \in \sR^2$ of the Bézier curves. Using DiffVG, a differentiable rasterizer $R$, by~\citet{li2020differentiable}, the vector form glyph is rasterizer into an image $\rvx = R(\gG)$ that allows for backpropagation through the image, \ie, $\frac{\partial \rvx}{\partial \vg_i}$ can be computed. This allows for the use of gradient-based methods for updating the glyph's control points. The choice of the objective function dictates how the glyph will turn out after optimization. We will next review how to use a pre-trained diffusion model as the objective.

\myparagraph{Diffusion models and score distillation.}
Text-to-image diffusion models~\cite{rombach2022high} aim to learn a conditional distribution $p(\rvx| \rvc)$ of the image $\rvx$ given the embedding $\rvc$ of a high-level concept described in natural language. The high-level idea behind a diffusion model is to learn to reverse the process of corrupting the input data with additive Gaussian noise at different levels of $\sigma$, \ie, learning a denoiser $D(\vx; \sigma)$. Prior works~\cite{hyvarinen2005estimation, song2019generative} have shown that
the denoiser can be interpreted as the score function, \ie, the gradient field of the data log-likelihood 
\bea\label{eq:score}
\nabla_\rvx \log p_\sigma(\rvx|\rvc) \approx (D(\rvx;\sigma)-\rvx)/\sigma^2.
\eea

Seminal works, DreamFusion~\cite{dreamfusion} and Score Jacobian Chanining (SJC)~\cite{wang2023score} proposed to distill the score function for the task of generating 3D assets. The high-level idea is to apply
 ``chain rule'' to the score function and backpropagate through a differentiable renderer to generate 3D assets. Given an image $\vx=R(\theta)$ that is rendered from 3D parameters $\theta$, the gradient of the conditional distribution \wrt $\theta$ is
 \bea
 \frac{\partial p_\sigma(\rvx|\rvc)}{\partial \theta} =  \underbracket{\frac{\partial p_\sigma(\rvx|\rvc)}{\partial \rvx}}_{\tt score} \frac{\partial \rvx}{\partial \theta}.
 \eea
 In other words, a pre-trained diffusion model can be used to update any image representation when provided with a differentiable renderer. In practice,~\citet{wang2023score} found that the rendered image $\rvx$ leads to an out-of-distribution issue as the images are not noisy but the denoiser is trained on noisy images. Hence, instead of~\equref{eq:score}, they propose to use the Perturb-and-Average Scoring (PAAS):
 \bea\label{eq:paas}
 \text{PAAS}(\rvx, \sigma) = \mathbb{E}_{\vn \sim \gN(0,\mI)} (D(\rvx + \sigma \vn;\sigma)-\rvx)/\sigma^2.
 \eea
 We note that~\equref{eq:paas} is equivalent to score distillation~\cite{dreamfusion}, albeit derived from a different mathematical perspective.

\section{Approach}
Our goal is to design ambigrams, \ie, a composition of glyphs that can be read from different orientations. While we describe our approach using rotational ambigrams, we note that the framework is easily generalizable to other types of ambigrams. The overview of the approach is shown in~\figref{fig:pipeline}.

\myparagraph{Problem formulation.}  We aim to construct an ambigram that reads to be the word $\va$ in the up-right orientation and read $\vb$ with rotated by 180 degrees. We denote rotation using the transformation $T$. We assume the two words $\va = (a_1,\dots,a_N)$ and $\vb=(b_1,\dots,b_N)$ have equal length $N$ with $a_n, b_n$ denoting letters in the English alphabet.

We formulate the design process of ambigrams as an optimization problem:
\bea\label{eq:all_obj}
\tG^\star = \argmin_{\tG} \gL_{\tt Ambi}(\tG, \va, \vb) + \gL_{\tt Style}(\tG),
\eea
where $\tG^\star = \left(\gG^{(1)}, \dots, \gG^{(N)}\right)$ denotes a sequence of  control point sets $\gG^{(n)}$ representing the $n^{\text{th}}$ glyph of the designed ambigram.

The objective consists of two terms: $L_{\tt Ambi}$ encourages the composition of glyphs form an ambigram and $\gL_{\tt Style}$ encourages the glyphs to have the style of a given font. We now describe each of the objectives in more detail.

\myparagraph{Ambigram loss.}
The ambigram loss %
is further decomposed into letter-level loss and word-level loss, \ie,
\bea\label{eq:all}
\gL_{\tt Ambi}(\tG, \va,\vb) = \gL_{\tt Letr}(\tG, \va,\vb) + \gL_{\tt Word}(\tG, \va,\vb).
\eea
The letter-level loss $\gL_{\tt Letr}$ encourages that each letter in the ambigram can be viewed from a different orientation. 
\bea\nonumber
\gL_{\tt Letr}(\tG, \va,\vb) = 
-\sum_{n=1}^N \Bigg[ \lambda_{\tt Letr}\log \left(p_{\sigma}(\vx^{(n)}|\vc(a_n))\right)\\
\mkern-18mu
+ (1-\lambda_{\tt Letr})\log \left(p_{\sigma}(T(\vx^{(n)})|\vc(b_{N-n+1}))\right)
\Bigg],
\label{eq:loss_letter}
\eea
where $p_\sigma(\vx|\vc)$ denotes the conditional probability from a pre-trained diffusion model, $\vx^{(n)}= R(\gG^{(n)})$ denotes the rasterized image using DiffVG~\cite{li2020differentiable}, $T$ denotes the rotation transformation, $\vc(a_n)$ corresponds to the embedding constructed using a text-prompt given the input letter $a_n$, and $\lambda_{\tt Letr} \in [0,1]$ denotes the weight balancing between the two orientations. Intuitively, the first term encourages the $n^{\text{th}}$ glyph to look like the letter $a_n$, and the second term encourages it to look like the letter $b_n$ when rotated.

Next, the word-level loss $\gL_{\tt word}$ aims for the \textit{entire word} to form an ambigram. To model dependencies between letters, the word-level loss is a total loss among all \textit{pairs} of consecutive letters,~\ie, $\gL_{\tt Word}(\tG, \va,\vb) = $
\bea\nonumber
-\sum_{n=1}^{N-1} \Bigg[\log \left(p_{\sigma}(\vx^{(n:n+1)}|\vc(a_n,a_{n+1}))\right)\\
+ \log \left(p_{\sigma}(T(\vx^{(n:n+1)})|\vc(b_{N-n},b_{N-n+1}))\right)
\Bigg],
\label{eq:loss_word}
\eea
where $\vx^{(n:n+1)} = [R(\tG^{(n)}), R(\tG^{(n+1)})]$ denotes the concatenation of the rasterized image for glyphs $\gG^{(n)}$ and $\gG^{(n+1)}$, $\vc(a_n,a_{n+1})$ corresponds to the embedding constructed using a text prompt given the two letters $a_n$ and $a_{n+1}$. This loss encourages pairs of glyphs to form an ambigram. 

\myparagraph{Style loss.} We propose to include additional losses to encourage different font styles. We propose the style loss $\gL_{\tt Style}(\gG)$ to control the style aspect of the glyphs that do not depend on the word letters. Specifically, the style loss consists of the font loss and a self-consistency loss:
\bea
\gL_{\tt Style}(\tG) = \left(\lambda_{\tt Font}\gL_{\tt Font}(\tG) + \lambda_{\tt Const}\gL_{\tt Const}(\tG)\right). 
\eea

\begin{figure}[t]
    \centering
    \setlength{\tabcolsep}{4pt}
    \renewcommand{\arraystretch}{0.85}
    \begin{tabular}{ccc|cc}
    \specialrule{.15em}{.05em}{.05em}
    & \multicolumn{2}{c|}{`B' $\leftrightarrow$ `D'} & \multicolumn{2}{c}{`C' $\leftrightarrow$ `E'}\\
    & Initial & Optimized & Initial & Optimized \\
    \hline
    \vspace{-0.1cm}
    Naive & 
        \raisebox{-.5\height}{\includegraphics[width=0.065\textwidth]{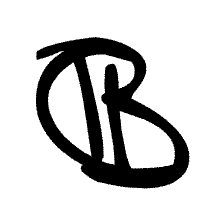}} &
        \raisebox{-.5\height}{\includegraphics[width=0.065\textwidth]{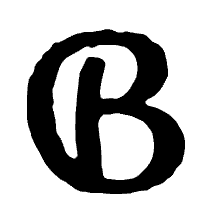}} &
        \raisebox{-.5\height}{\includegraphics[width=0.065\textwidth]{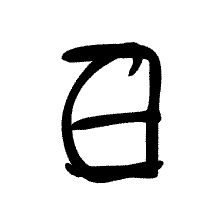}} &
        \raisebox{-.5\height}{\includegraphics[width=0.065\textwidth]{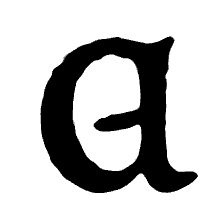}} \\
    Max Overlap & 
        \raisebox{-.5\height}{\includegraphics[width=0.065\textwidth]{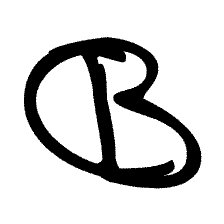}} &
        \raisebox{-.5\height}{\includegraphics[width=0.065\textwidth]{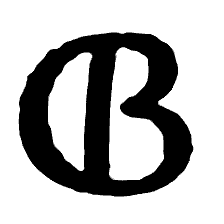}} &
        \raisebox{-.5\height}{\includegraphics[width=0.065\textwidth]{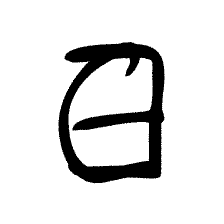}} &
        \raisebox{-.5\height}{\includegraphics[width=0.065\textwidth]{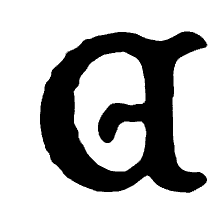}} \\
    Contact (L) & 
        \raisebox{-.5\height}{\includegraphics[width=0.065\textwidth]{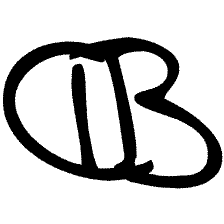}} &
        \raisebox{-.5\height}{\includegraphics[width=0.065\textwidth]{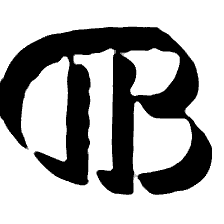}} &
        \raisebox{-.5\height}{\includegraphics[width=0.065\textwidth]{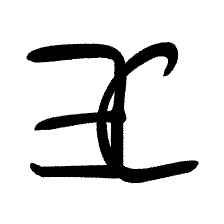}} &
        \raisebox{-.5\height}{\includegraphics[width=0.065\textwidth]{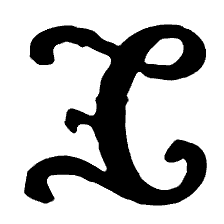}} \\
    Contact (R) & 
        \raisebox{-.5\height}{\includegraphics[width=0.065\textwidth]{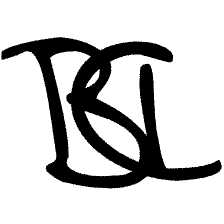}} &
        \raisebox{-.5\height}{\includegraphics[width=0.065\textwidth]{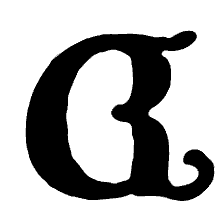}} &
        \raisebox{-.5\height}{\includegraphics[width=0.065\textwidth]{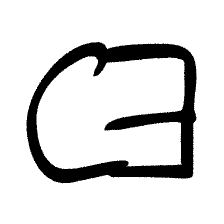}} &
        \raisebox{-.5\height}{\includegraphics[width=0.065\textwidth]{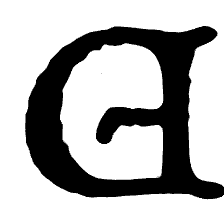}} \\
        \specialrule{.15em}{.05em}{.05em}
    \end{tabular}
    \vspace{-0.25cm}
    \caption{Illustration of the proposed alignment strategies. We show the effect of each proposed initialize scheme before and after optimization. Observe that, depending on the initialization the design varies.
    }
    \vspace{-0.2cm}
    \label{fig:alignment_strategy}
\end{figure}
The font loss $\gL_{\tt Font}(\tG)$ encourages the generated glyph to be closer.
We use a trained font attribute predictor $A$ by~\citet{wang2020attribute2font} and minimize the $\ell_2$ difference between the predicted attribute for all glyphs $\gG^{(n)}$ and the attribute vectors $v$ of a chosen font. Formally,
\bea
\gL_{\tt font}(\tG) = \sum_{n=1}^N \mathbb{E}_{v}\norm{v-A(R(\gG^{(n)}))}_2^2,
\eea
where $v$ is uniformly sampled from the set of attribute vectors of a desirable font extracted using the same attribute predictor $A$. 

Next, for ambigrams where $\va$ and $\vb$ are the same words, we further impose a self-consistent loss $\gL_{\tt Const}$ encourages self-similarity after the transformation, \ie,
\bea
\sum_{n} \norm{\texttt{Blur}(\vx^{(N-n)}) - \texttt{Blur}(T(\vx^{(n)})))}_2^2,
\eea
where $\vx^{(n)} = R(\gG^{(n)})$.
We perform Blurring with a $3\times 3$ Gaussian filter
to ensure the overall shape matches.

\myparagraph{Gradient-based optimization.} To solve the minimization program in~\equref{eq:all}, we use the Adam~\cite{adam} optimizer to update the Berizer curves' control points. For computing the gradient through the diffusion model, we use PAAS as reviewed in~\equref{eq:paas}. In practice, we perform stage-wise optimization. We first optimize for individual letters, then we jointly tune all letters together using the word loss. We now discuss additional algorithmic details when performing the letter-level and word-level optimization, including initialization, hyperparameters, and post-processing procedures.

\subsection{Letter-level optimization}\label{sec:letter}
\myparagraph{Glyph initialization and alignment.} To solve the optimization in~\equref{eq:all_obj}, we need to initialize the control points for the glyphs $\tG=(\gG^{(1)}, \dots \gG^{(N)})$. We propose to initialize the control points by overlaying existing fonts. Given the words $\va = (a_1,\dots,a_N)$ and $\vb=(b_1,\dots,b_N)$, the glyph $\gG^{(n)}$ will be initialized with the control points from a pre-defined font of $a_n$ and $b_{N-n+1}$ but rotated. 

One of the key challenges is how to align the two existing fonts. Different alignment generates very different designs after optimization. We propose to align the letters using horizontal and vertical shifts based on four different schemes %
\begin{itemize}
    \item {\it Naive}: Directly overlap the two letters.
    \item {\it Max Overlap}: We align the two letters to have the maximum number of overlapping pixels.
    \item {\it Contact (Left)}: We find the leftmost shift such that the two letters are still in contact, and then we multiply this shift amount by 0.7 to ensure the letters overlap.
    \item {\it Contact (Right)}: The same as left contact, but for the rightmost shift.
\end{itemize}
The effects of these alignment schemes, before and after optimization, are visualized in~\figref{fig:alignment_strategy}. We observe that different alignment strategies lead to different designs that would otherwise be difficult to generate with another initialization scheme. The choice of the alignment scheme is treated as a hyperparameter.

\begin{figure}[t]
    \centering
    \setlength{\tabcolsep}{4pt}
    \begin{tabular}{ccccc}
    \specialrule{.15em}{.05em}{.05em}
    & `A` $\leftrightarrow$ `g` & `C` $\leftrightarrow$ `W` & `S` $\leftrightarrow$ `Z` & `B` $\leftrightarrow$ `B`\\
    \hline
    Before 
    & \raisebox{-.35\height}{\includegraphics[height=1.6cm]{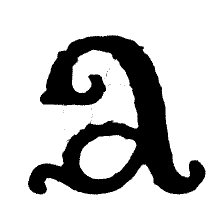}}
    & \raisebox{-.35\height}{\includegraphics[height=1.6cm]{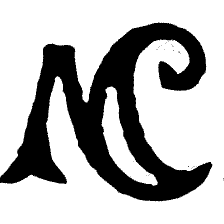}}
    & \raisebox{-.35\height}{\includegraphics[height=1.6cm]{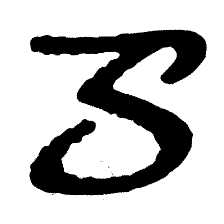}}
    & \raisebox{-.35\height}{\includegraphics[height=1.6cm]{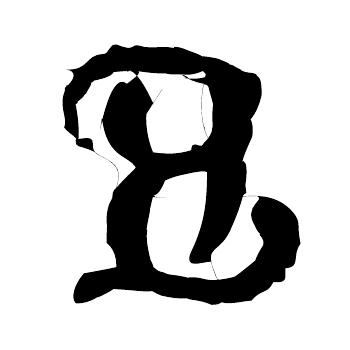}}
    \\
    After 
    & \raisebox{-.35\height}{\includegraphics[height=1.6cm]{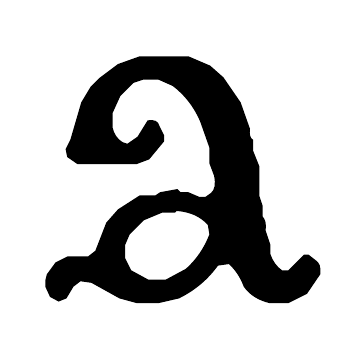}}
    & \raisebox{-.35\height}{\includegraphics[height=1.6cm]{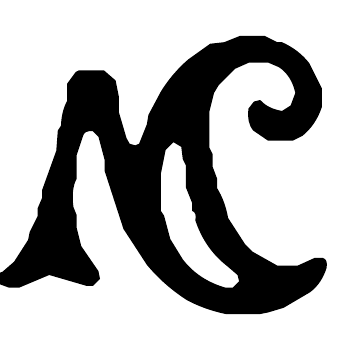}}
    & \raisebox{-.35\height}{\includegraphics[height=1.6cm]{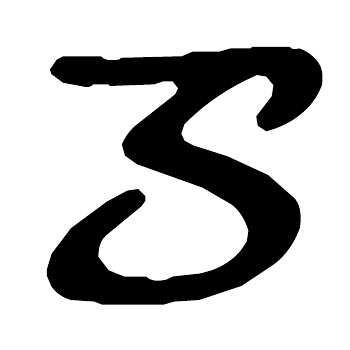}}
    & \raisebox{-.35\height}{\includegraphics[height=1.6cm]{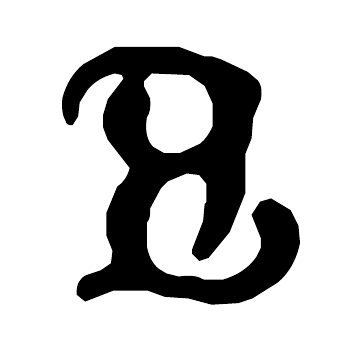}}\\
    \specialrule{.15em}{.05em}{.05em}
    \end{tabular}
    \vspace{-0.25cm}
    \caption{Illustration of before and after post-processing of median filter and image sharpening. Observe the reduction in floaters and smoother contours.
    }
    \vspace{-0.25cm}
    \label{fig:post_process}
\end{figure}
\begin{figure*}[t]
\small
\centering
\setlength{\tabcolsep}{5pt}
\renewcommand{\arraystretch}{1.8}
\begin{tabular}{ccccc}
Method & ``THE'' & ``AREA'' & ``AMONG'' & ``BEAUTY''\\
\raisebox{6pt}{Ambigramania} & 
\includegraphics[height=0.68cm]{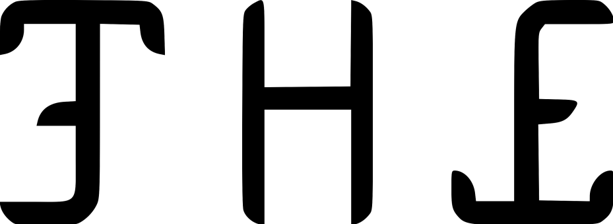} &
{\includegraphics[height=0.68cm]{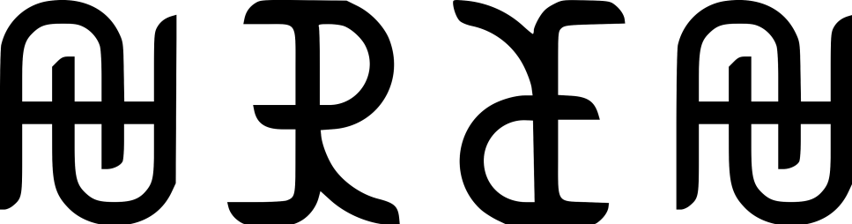}} &
\includegraphics[height=0.68cm]{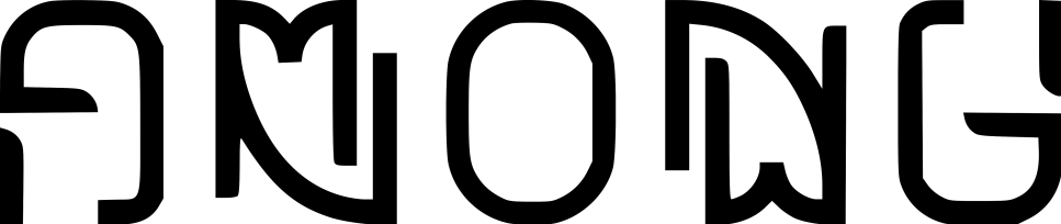} &
\includegraphics[height=0.68cm]{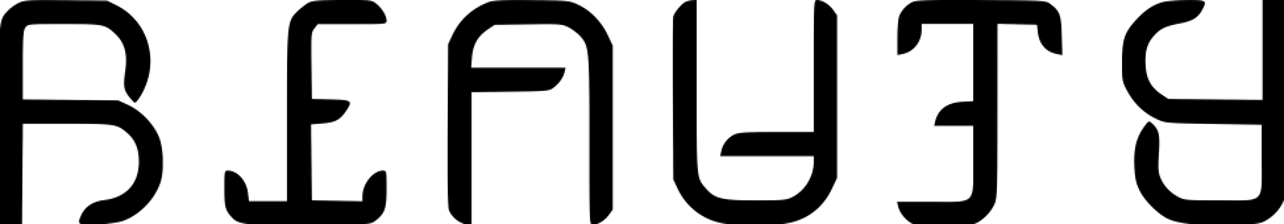}\\
\raisebox{6pt}{Ambimaticv2} &
{\includegraphics[height=0.68cm]{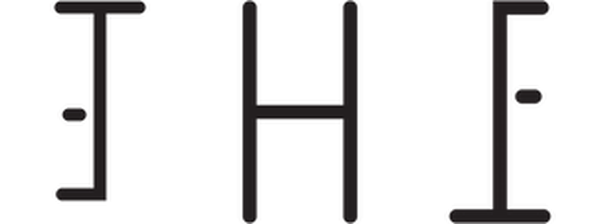}} &
{\includegraphics[trim={1cm 0.2cm 1cm 0.2cm},clip, height=0.68cm]{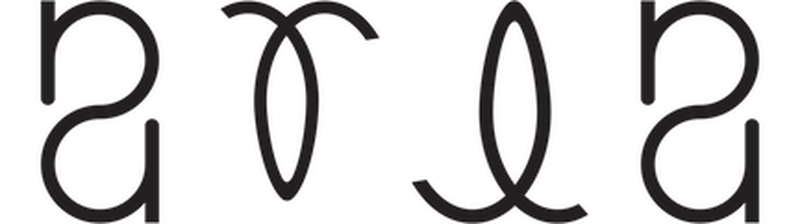}} &
{\includegraphics[trim={1cm 0.2cm 1cm 0.2cm},clip,height=0.68cm, width=3.4cm]{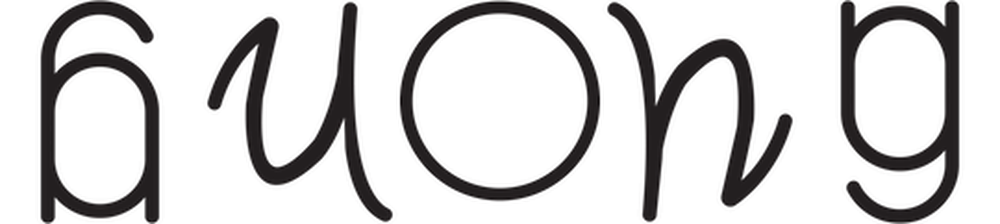}} &
{\includegraphics[trim={1cm 0.2cm 1cm 0.2cm},clip, height=0.68cm, width=3.8cm]{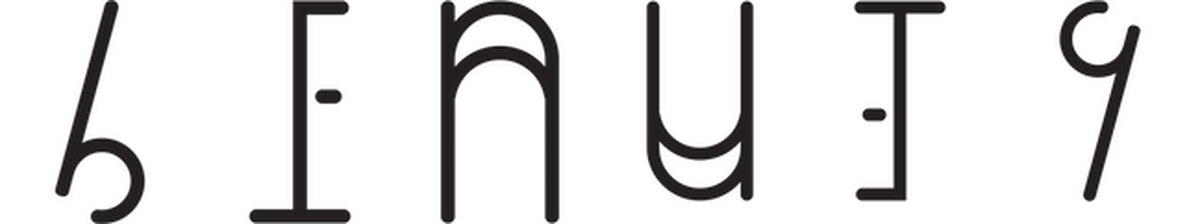}}\\
\raisebox{6pt}{DsmonoHD} &
\includegraphics[height=0.68cm]{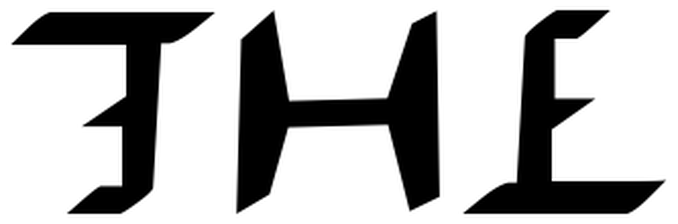} &
{\includegraphics[height=0.68cm]{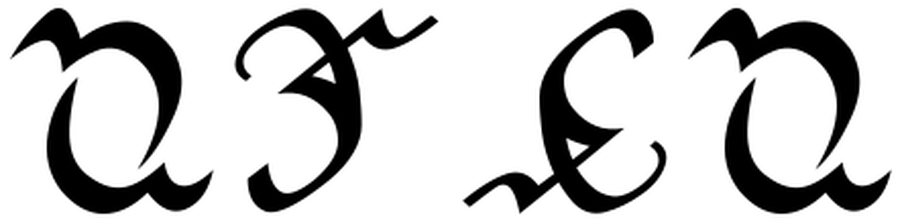}} &
\includegraphics[height=0.68cm]{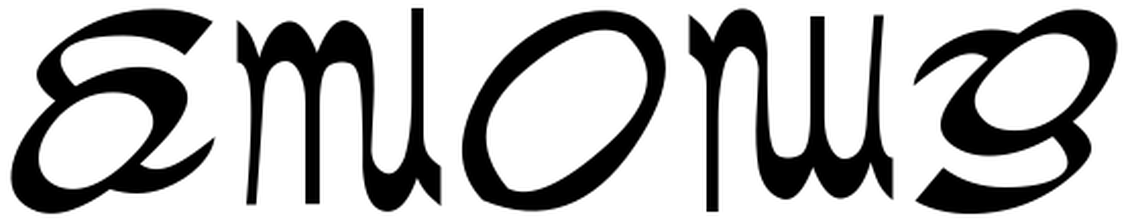} &
\includegraphics[height=0.68cm]{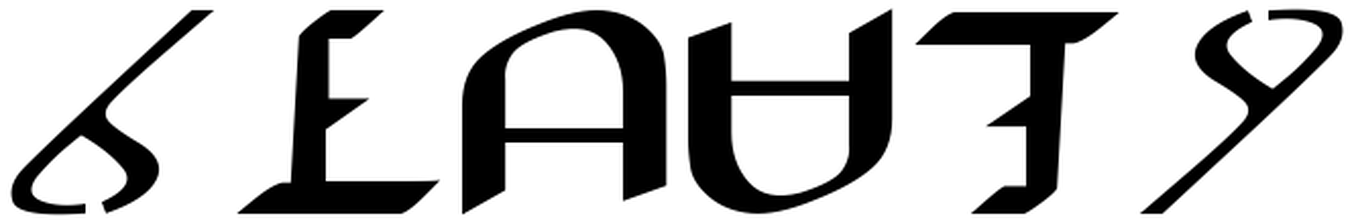}\\
\noalign{\vskip -0.05cm}
\hline\noalign{\vskip 0.15cm}
\raisebox{6pt}{Ambidream} & 
{\includegraphics[trim={1cm 0.4cm 1cm 0.4cm},clip,height=0.68cm]{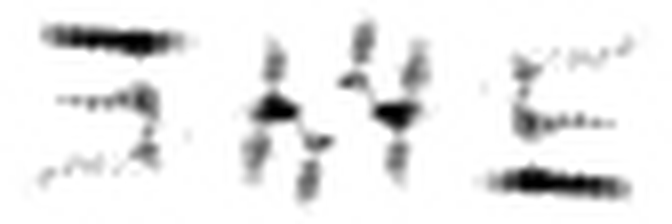}} &
{\includegraphics[trim={1cm 0.4cm 1cm 0.4cm},clip,height=0.68cm]{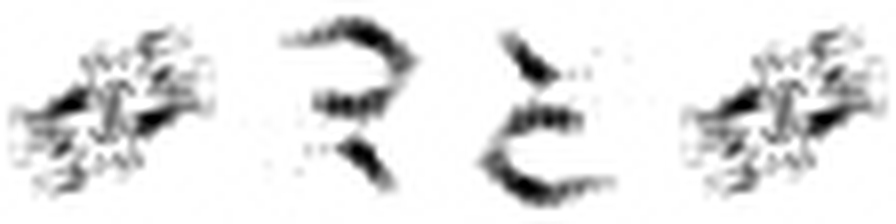}} &
{\includegraphics[trim={1cm 0.4cm 1cm 0.4cm},clip,height=0.68cm]{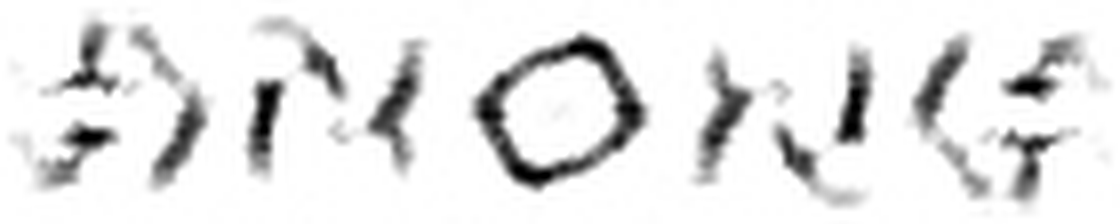}} &
{\includegraphics[trim={1cm 0.4cm 1cm 0.4cm},clip,height=0.68cm]{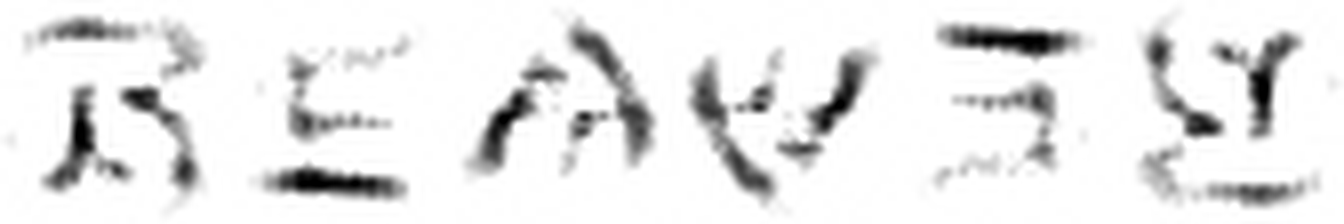}}\\
\raisebox{6pt}{Ambifusion} & 
{\includegraphics[trim={0.4cm 1cm 0.4cm 1cm},clip,height=0.68cm, width=2.3cm]{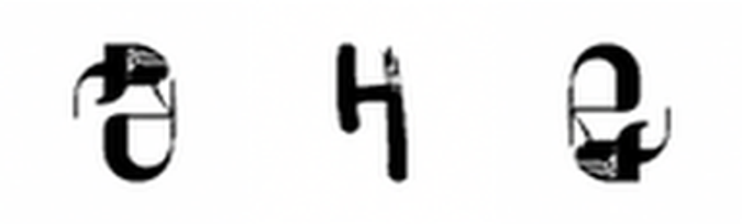}} &
{\includegraphics[trim={0cm 1cm 0cm 1cm},clip,height=0.68cm, width=3.5cm]{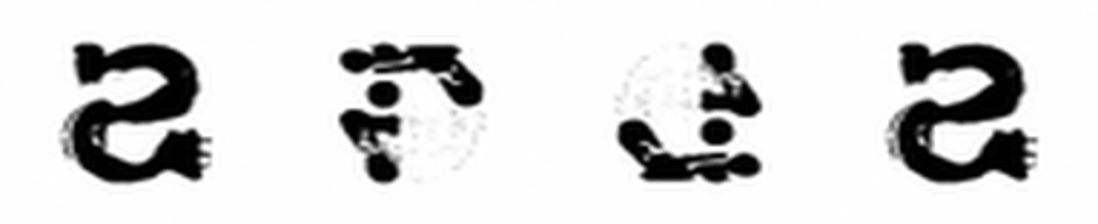}} &
{\includegraphics[trim={0cm 1cm 0cm 1cm},clip,height=0.68cm, width=3.8cm]{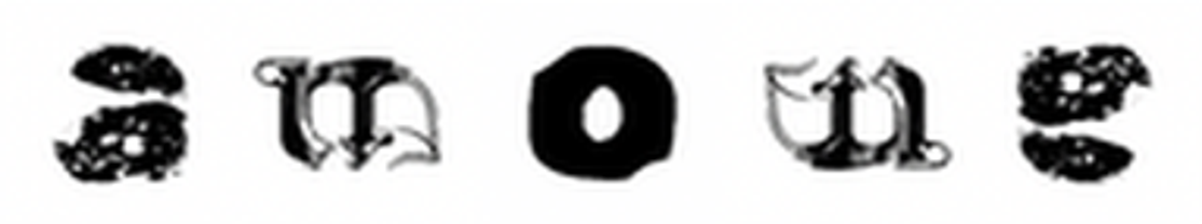}} &
{\includegraphics[trim={0cm 1cm 0cm 1cm},clip,height=0.68cm, width=4.3cm]{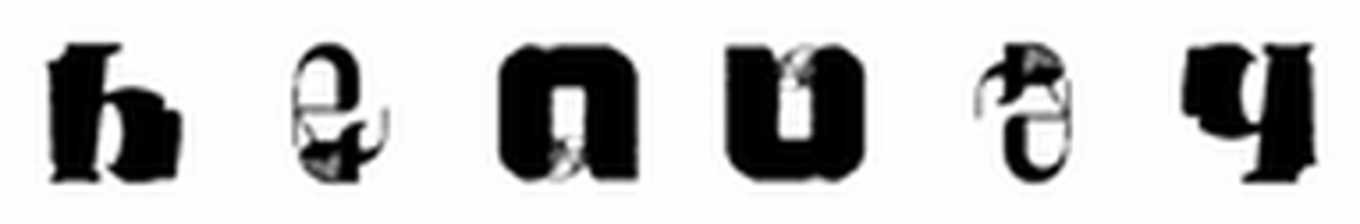}}\\
\raisebox{6pt}{Ours (Font 1)} & 
\includegraphics[height=0.68cm]{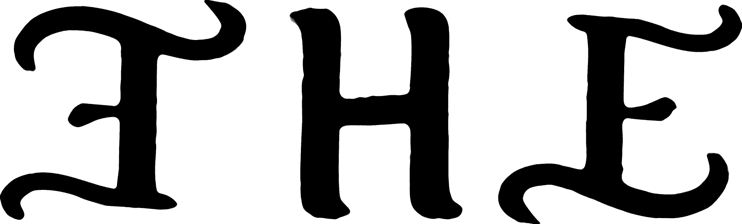} &
\includegraphics[height=0.68cm]{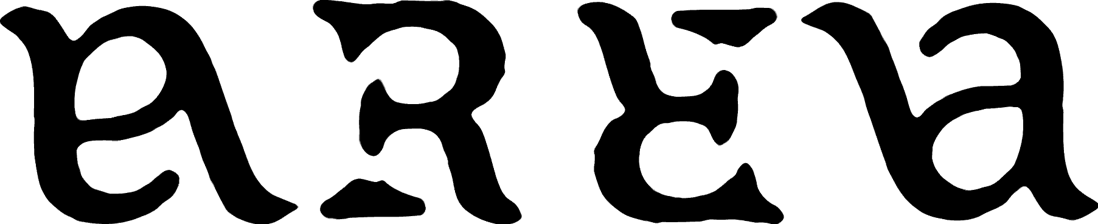} &
\includegraphics[height=0.68cm]{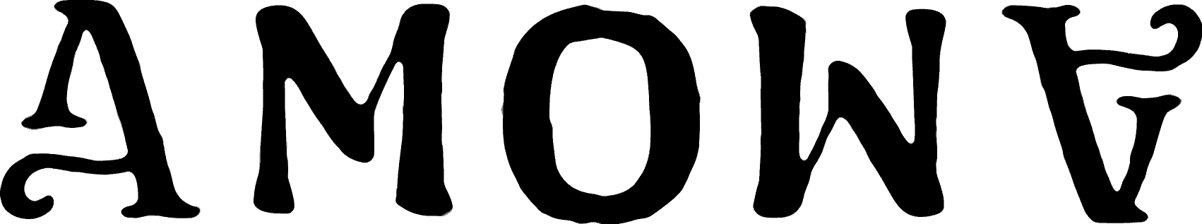} &
\includegraphics[height=0.68cm]{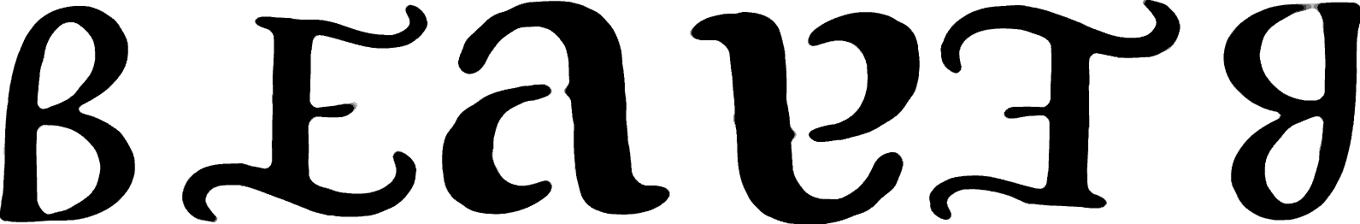}\\
\raisebox{6pt}{Ours (Font 2)} & 
\includegraphics[height=0.68cm]{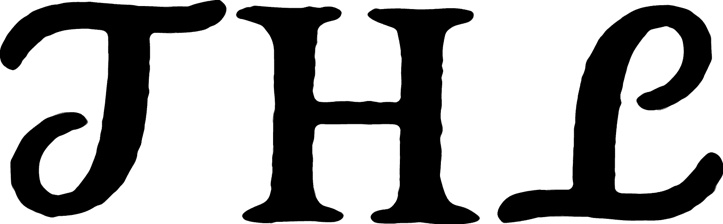} &
\includegraphics[height=0.68cm]{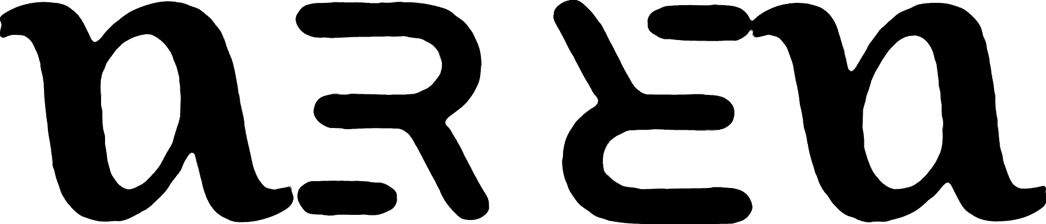} &
\includegraphics[height=0.68cm]{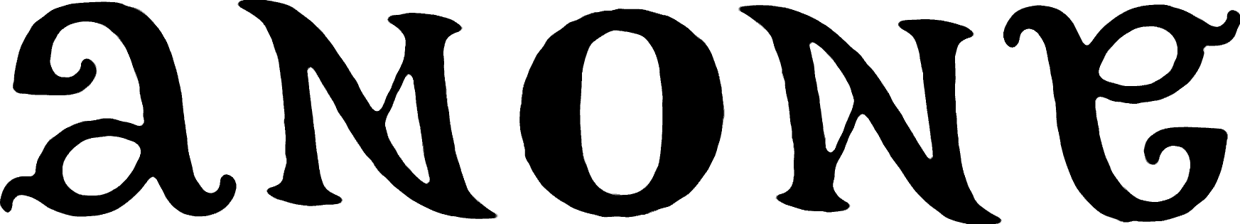} &
\includegraphics[height=0.68cm]{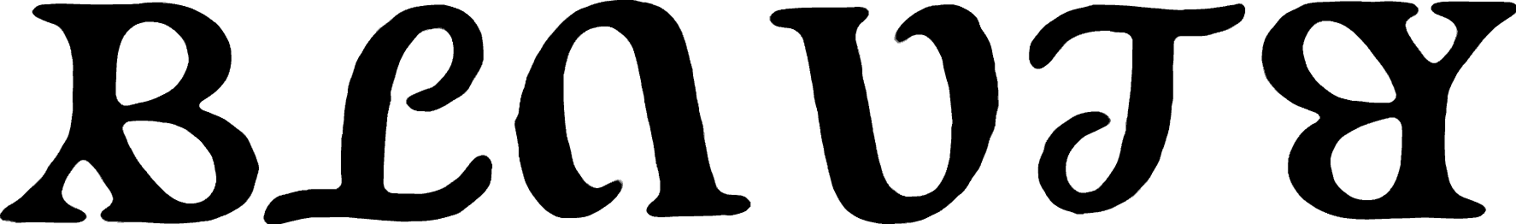} \\

\end{tabular}
\vspace{-0.25cm}
\caption{Qualitative comparison across baselines. Methods above the horizontal line are designed by artists. The generations of Ambigramania are from~\url{ambigramania.com}, Ambimaticv2, DsmonoHD, and Ambidream are from~\url{makeambigrams.com}. Ambifusion's samples are generated using the official code from~\url{github.com/univ-esuty/ambifusion}. 
}
\vspace{-0.2cm}
\label{fig:comparison}
\end{figure*}
\myparagraph{Optimization details.}
To compute the gradient through the diffusion model, we need to provide the embedding $\vc(a_n)$ and $\vc(b_{N-n+1})$. We use the text prompt of \texttt{``An image of the letter \{\} in lower/upper case.''} where $\{\}$ is replaced with the letter of $a_n$ and $b_{N-n+1}$.

Next, we perform random perspective as data augmentation on the rasterized image. The augmented images form a batch size of five for which we perform one gradient update step. %
In total, we perform 500 gradient steps using the Adam optimizer~\cite{adam} with an exponentially decayed learning rate. We also decay the style weight $\lambda_{\tt Style}$ exponentially.

\myparagraph{Automatic design selection and post-processing.} 
The proposed optimization involves several hyperparameters, \eg, the initialize scheme and the $\lambda$s to weigh the losses. Instead of manually inspecting all the generated designs, we rank the promising designs by sorting based on the cross-entropy loss of a trained character classifier; see appendix for details.
The classifier judges whether the design can be viewed correctly as the given letter, and thus can quickly filter out unpromising designs. To form an ambigram font, we select $26\times26$ %
design one for each pair of letters in the alphabet. 

Next, while the optimization procedure generates a promising design with correct shapes, the edges may be jagged and contain thin artifacts. To have a more robust selection, we use image processing techniques as post-processing to eliminate floaters in the generated glyph. This includes a sequence of median filters and image sharpening on the rasterized image. %
See~\figref{fig:post_process} where we visualize the before and after post-processing. Overall, the post-processed results exhibit a smoother and aesthetically more pleasing glyph. 
We will next describe how we can further improve the design by considering word-level semantics.

\subsection{Word-level optimization}\label{sec:word}
With the $N$ individual glyphs designed, we perform joint optimization over pairs of letters as formulated in~\equref{eq:all}. To do so, we would need to align the designed glyphs.%

\myparagraph{Word initialization and alignment.}
Given the designed sequence of glyphs from~\secref{sec:letter}, we would need to scale and align them to form an ambigram. To align the individual letters into a word, we created a template consisting of $N$ equally spaced rectangles. Next, we center each of the glyphs inside the rectangle. We then linearly scale the glyph until its height or width matches the rectangle while maintaining the aspect ratio. This sequence of glyphs is used as the initialization to perform word-level optimization, which we describe next.

\myparagraph{Optimization details.} %
To compute the gradient through the diffusion model, we need to specify the embedding $\vc(\va)$ and $\vc(\vb)$. For this, we use the text prompt of ``\texttt{A blank paper with the text ``\{\}'' written on it}.'' where $\{\}$ is replaced with a pair of letters, \eg, $a_n,a_{n+1}$.
As in the letter-level stage, we perform random perspective with a distortion scale of 0.2 on the paired letter and optimize the Adam optimizer for 110 steps. Additionally, we introduce a regularization that prevents the glyphs from deviating too much from the designs from the letter-level optimization; see appendix for details.

\begin{figure*}[t]
\small
\centering
\setlength{\tabcolsep}{5pt}
\renewcommand{\arraystretch}{1.8}
\begin{tabular}{ccccc}
Method & ``AND'' & ``BASE'' & ``DRIVE'' & ``CORRECT''\\
\raisebox{6pt}{Ambigramania} & 
\includegraphics[height=0.65cm]{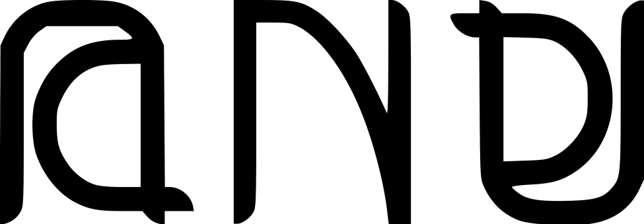} &
\includegraphics[height=0.65cm]{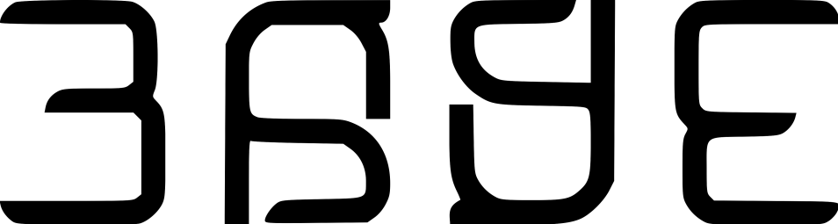} &
\includegraphics[height=0.65cm]{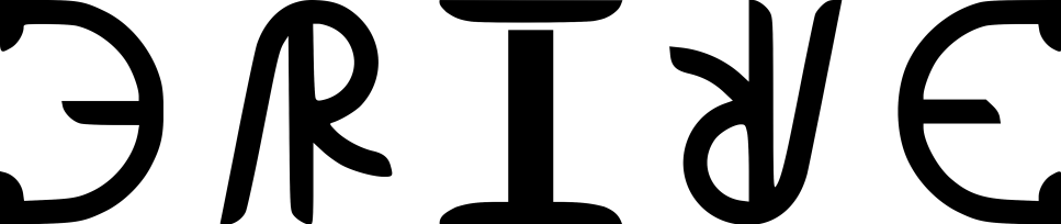} &
\includegraphics[height=0.65cm]{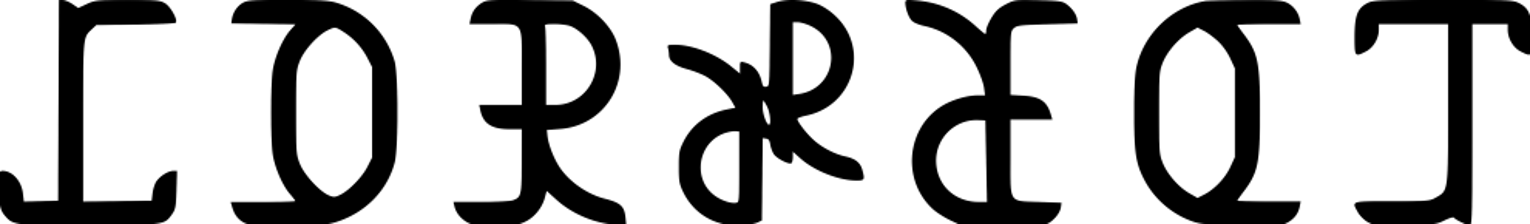}\\
\raisebox{6pt}{Ambimaticv2} &
{\includegraphics[trim={1cm 0.2cm 1cm 0.2cm},clip,height=0.65cm]{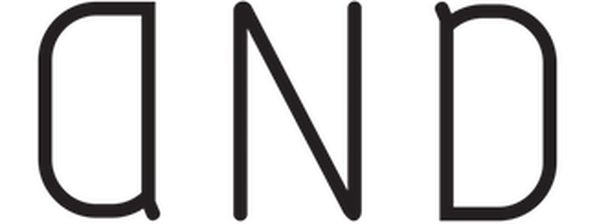}} &
{\includegraphics[trim={1cm 0.2cm 1cm 0.2cm},clip,height=0.65cm, width=2.7cm]{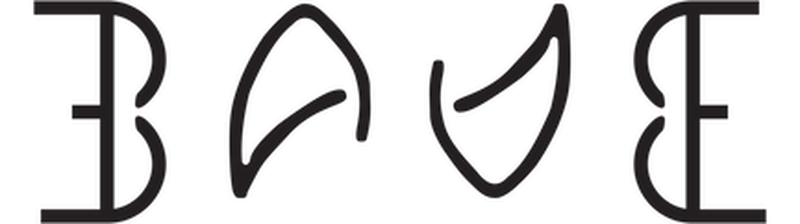}} &
{\includegraphics[trim={1cm 0.2cm 1cm 0.2cm},clip,height=0.65cm, width=3.2cm]{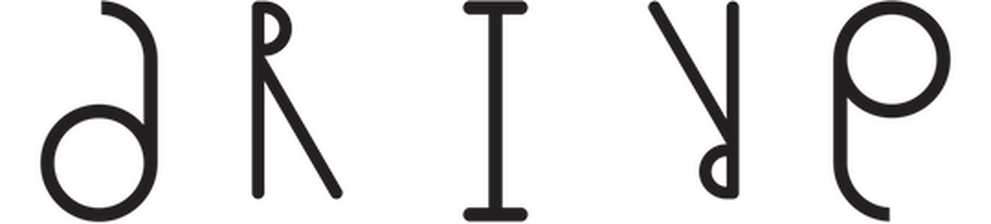}} &
{\includegraphics[trim={1cm 0.2cm 1cm 0.2cm},clip,height=0.65cm, width=4.3cm]{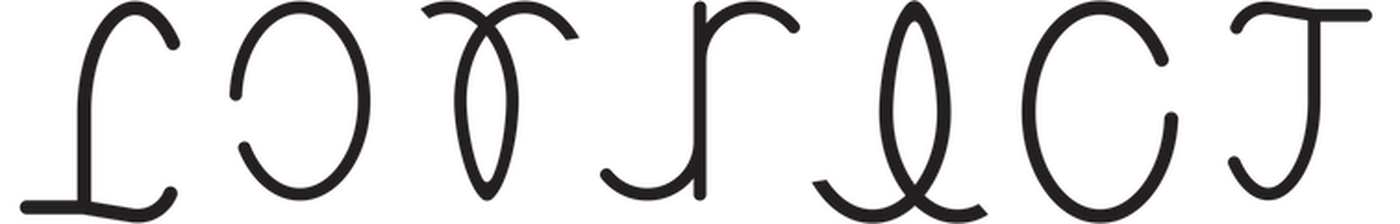}}\\
\raisebox{6pt}{DsmonoHD} &
\includegraphics[height=0.65cm]{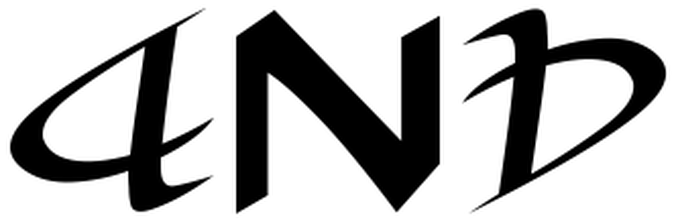} &
\includegraphics[height=0.65cm]{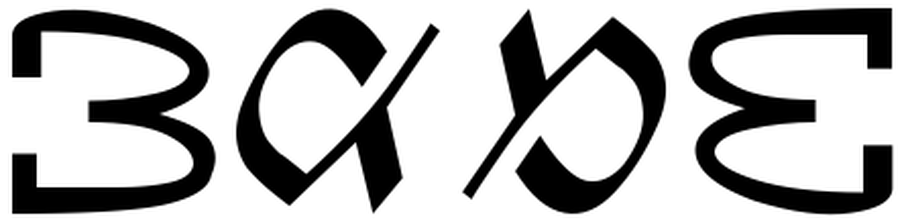} &
\includegraphics[height=0.65cm]{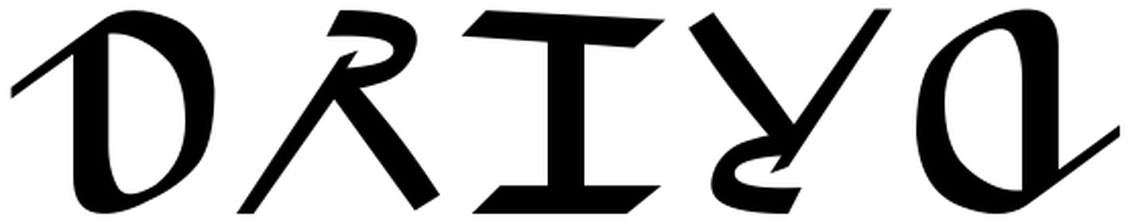} &
\includegraphics[height=0.65cm]{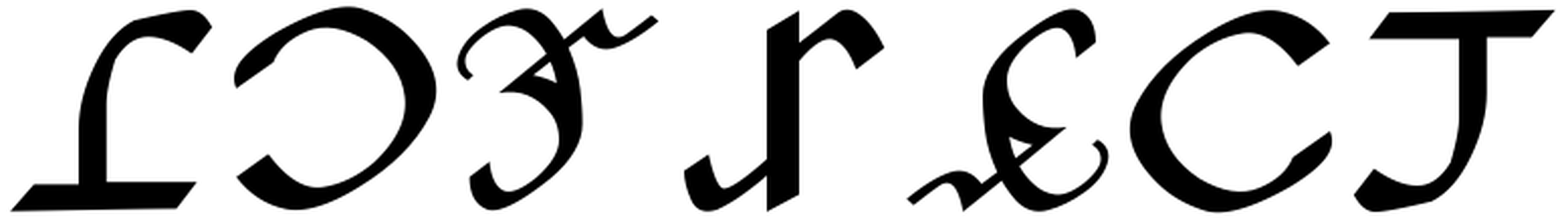}\\
\noalign{\vskip -0.05cm}
\hline\noalign{\vskip 0.15cm}
\raisebox{6pt}{Ambidream} & 
{\includegraphics[trim={1cm 0.4cm 1cm 0.4cm},clip,height=0.65cm]{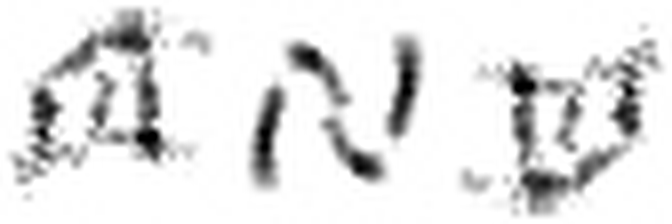}} &
{\includegraphics[trim={1cm 0.4cm 1cm 0.4cm},clip,height=0.65cm]{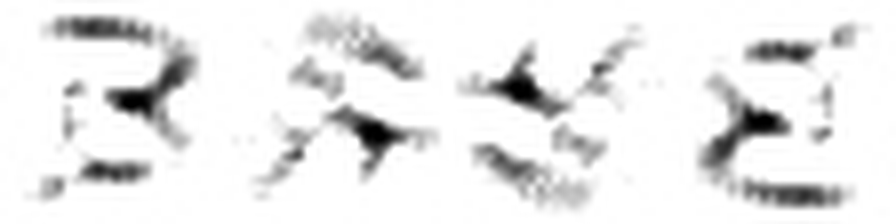}} &
{\includegraphics[trim={1cm 0.4cm 1cm 0.4cm},clip,height=0.65cm]{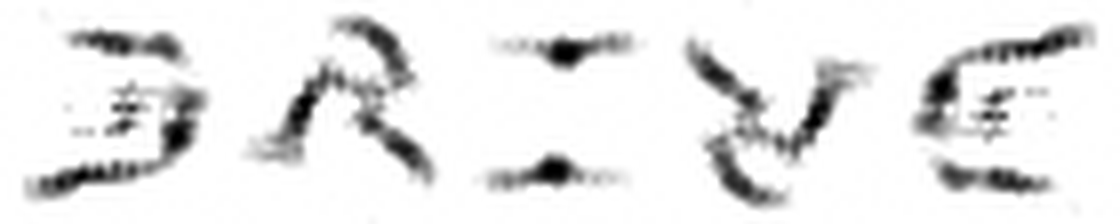}} &
{\includegraphics[trim={1cm 0.4cm 1cm 0.4cm},clip,height=0.65cm]{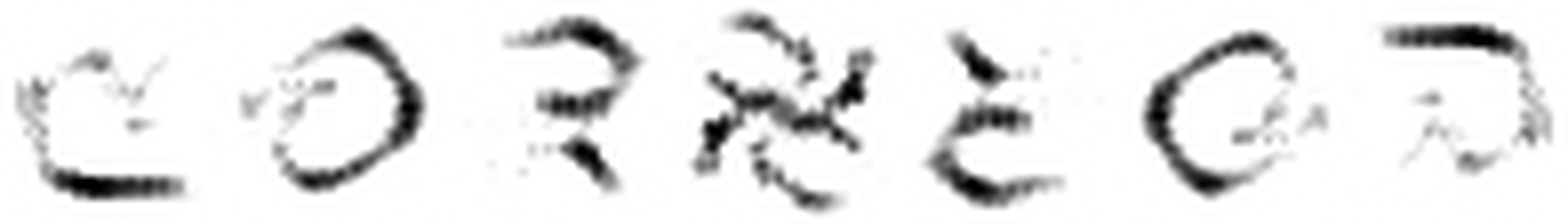}}\\
\raisebox{6pt}{Ambifusion} & 
{\includegraphics[trim={0.4cm 1cm 0.4cm 1cm},clip,height=0.65cm, width=2.3cm]{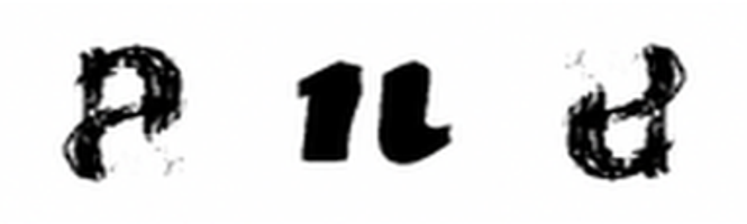}} &
{\includegraphics[trim={0.4cm 1cm 0.4cm 1cm},clip,height=0.65cm, width=3cm]{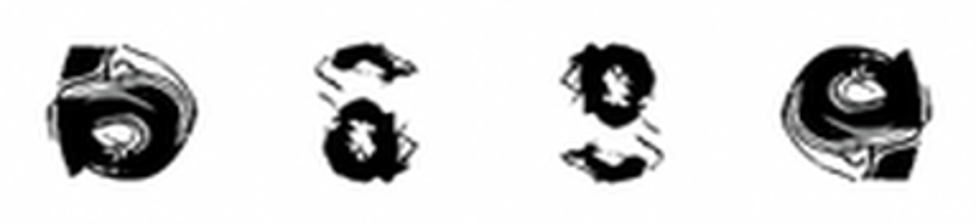}} &
{\includegraphics[trim={0.4cm 1cm 0.4cm 1cm},clip,height=0.65cm, width=3.8cm]{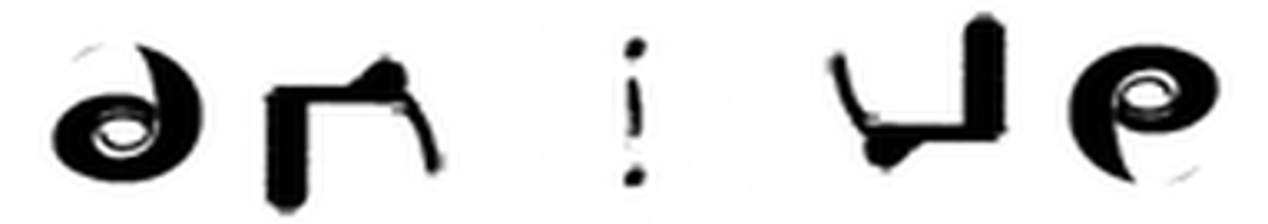}} &
{\includegraphics[trim={0.4cm 1cm 0.4cm 1cm},clip,height=0.65cm, width=4.9cm]{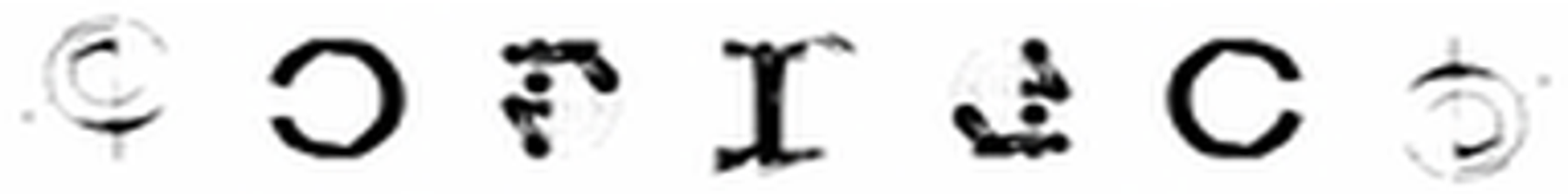}}\\
\raisebox{6pt}{Ours (Font 1)} & 
\includegraphics[height=0.65cm]{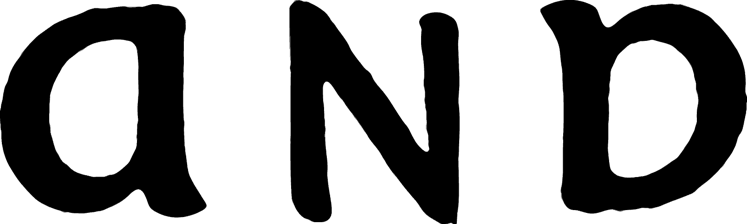} &
\includegraphics[height=0.65cm]{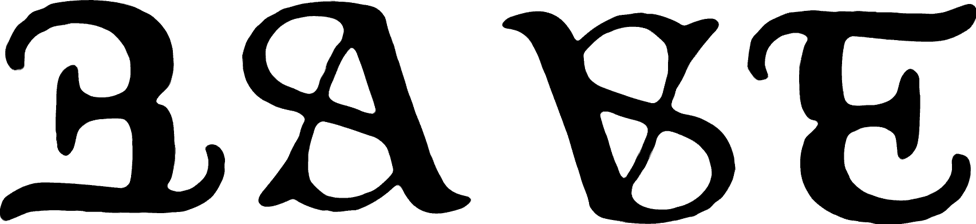} &
\includegraphics[height=0.65cm]{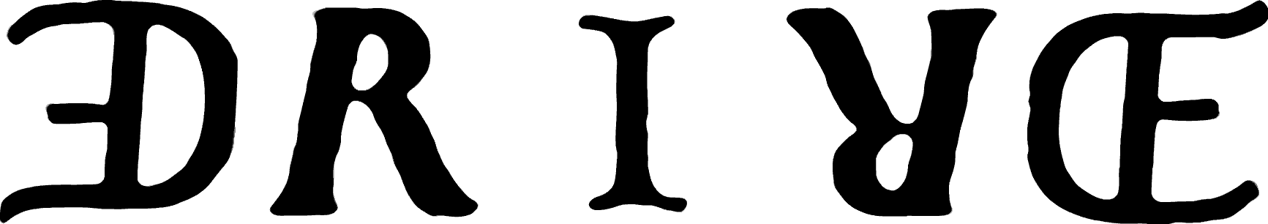} &
\includegraphics[height=0.65cm]{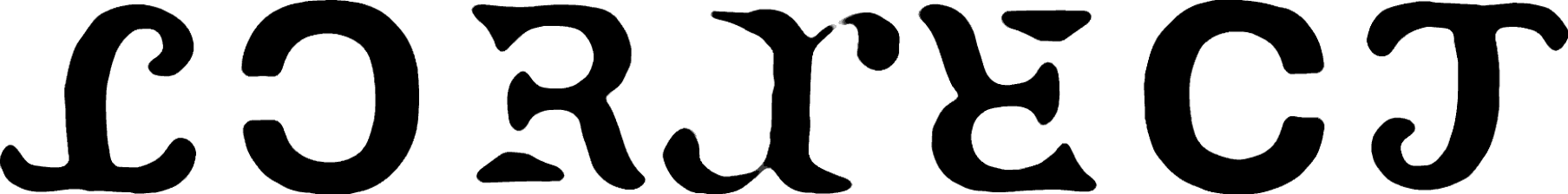}\\
\raisebox{6pt}{Ours (Font 2)} & 
\includegraphics[height=0.65cm]{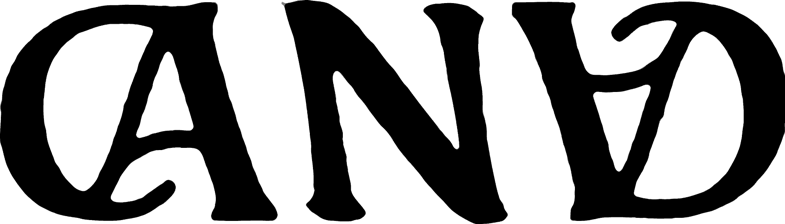} &
\includegraphics[height=0.65cm]{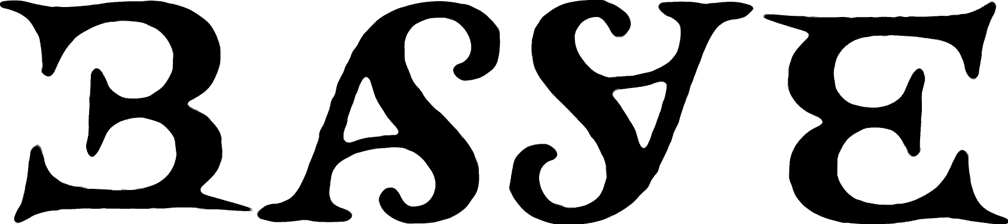} &
\includegraphics[height=0.65cm]{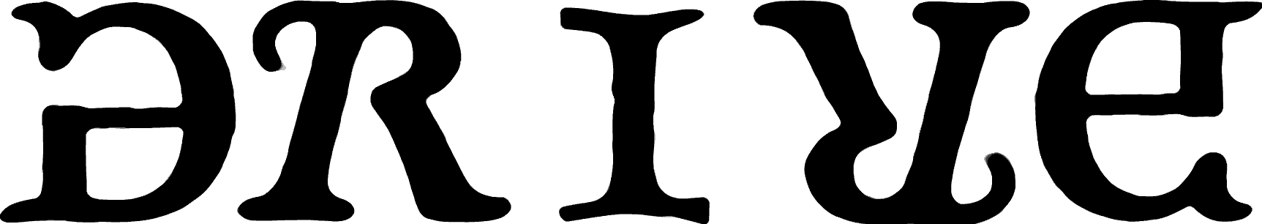} &
\includegraphics[height=0.65cm]{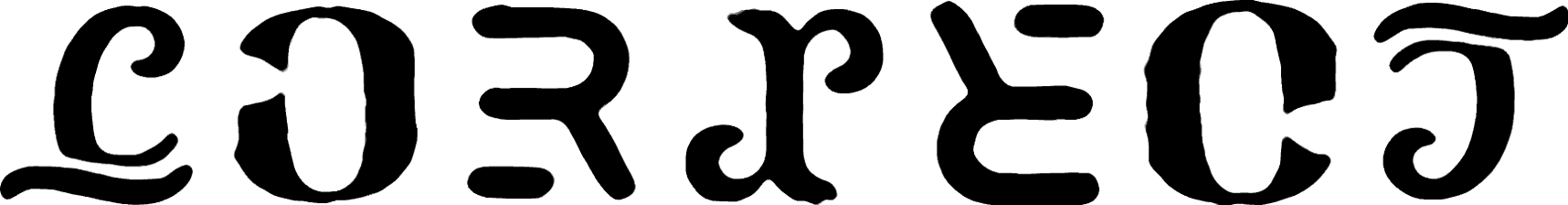} \\
\end{tabular}
\vspace{-0.25cm}
\caption{Additional qualitative comparison across baselines. Methods above the horizontal line are designed by artists. 
We observe that generations from Ours and DsmonoHD are the most legible while Ambidream's generation is the most difficult to read.
}
\vspace{-0.2cm}
\label{fig:comparison2}
\end{figure*}

\section{Experiments}
We conduct both quantitative and qualitative studies comparing existing methods for generating ambigrams. 
We report quantitative metrics for evaluating the legibility of the ambigrams in both orientations and show ample qualitative results. Finally, we conduct detailed ablation studies to analyze the design choices in our method, demonstrating the necessity of each component.

\myparagraph{Baselines.} For baselines, we consider five ambigram fonts, including DsmonoHD, Ambimaticv2, Ambigramania, Ambidream, and Ambifusion. The first three fonts were manually designed by artists obtained from online generation sites~\cite{ambigramania,makeambigram}.
Next, Ambidream~\cite{makeambigram} and Ambifusion~\cite{shirakawa2023ambigram} were generated using AI techniques as reviewed in~\secref{sec:rel}. For Ambifusion, we use the code released by~\citet{shirakawa2023ambigram}. We note that Ambifusion only generates rotational ambigram of a single letter directly in the pixel space. To generate an ambigram with multiple letters, %
we horizontally concatenated the individual letter images generated from their method.

\begin{table}[t]
\centering
\begin{tabular}{lcc}
\specialrule{.15em}{.05em}{.05em}
Method & Accuracy$\uparrow$ & Edit Distance$\downarrow$ \\
\midrule
Ambigramania~\cite{ambigramania}  & 6.2\% & 8.112 \\
Ambimaticv2~\cite{makeambigram} & 7.2\% & 7.670 \\
DsmonoHD~\cite{makeambigram}  & 9.2\% & 7.042 \\
\hline
Ambidream~\cite{ambidream} & 6.8\% & 7.212 \\
Ambifusion~\cite{shirakawa2023ambigram} & 3.6\% & 8.726 \\
Ours (Font 1) & \textbf{24.2\%} & \textbf{3.634} \\
Ours (Font 2) & 20.8\% & 4.088 \\
\specialrule{.15em}{.05em}{.05em}
\end{tabular}
\vspace{-0.2cm}
\caption{
{Quantitative results.} Methods above the horizontal line are designed by artists. Both of our fonts outperform baseline methods in accuracy and edit distance metrics.
}
\vspace{-0.25cm}
\label{tab:OCR_metric}
\end{table}
\myparagraph{Experiment setup.}
 For evaluation, we consider generating rotational ambigrams with $\va=\vb$,~\ie, the ambigram should read the same under a 180-degree rotation. We chose this setting as it is supported by all the aforementioned baselines.

To construct a benchmark, we use the 500 most common words in English that are longer than two characters~\cite{1000CommonWords}. As our method can generate multiple designs, we generated two sets of ambigram fonts for evaluation.

\subsection{Quantitative comparisons}

\myparagraph{Evaluation metrics.}
To evaluate the legibility of the generated words, we use TrOCR\cite{trocr} a transformer-based Optical Character Recognition (OCR) model to see if it can correctly recognize the generated ambigrams. We report two evaluation metrics:
\begin{itemize}
    \item {\it Accuracy$\uparrow$:} An ambigram is considered ``correct'', if TrOCR recognizes all letters in the word correctly for \textbf{both viewing orientations}.
    \item {\it Edit Distance$\downarrow$:}  We consider the Levenshtein edit distance between the ground truth and the predicted word from TrOCR summed over both orientations averaged over the dataset. Levenshtein distance counts the number of insertions, substitutions, and deletions to make the two strings identical.   
\end{itemize}

To our knowledge, we are the first to benchmark ambigrams at the word-level. Prior work~\cite{shirakawa2023ambigram} only considers single letter accuracy by training a ResNet~\cite{he2016deep} classifier on the MyFonts dataset~\cite{chen2019large}. We believe using TrOCR~\cite{trocr} and evaluating ambigrams at the word-level is more appropriate as it considers the relationship between the letters. We chose TrOCR as it has open-sourced the code and released the model weights. 

\myparagraph{Results.}
We report the quantitative results in~\tabref{tab:OCR_metric}. We observe that both of the fonts for our approach achieved better results with the highest accuracy of 24.2\% and the lowest edit distance of 3.634 out of all the methods. The best baseline, DsmonoHD (a font designed by an artist), achieved an accuracy of 9.2\% and an edit distance of 7.0. The other two AI-based methods achieved 6.8\% and 3.6\% in accuracy which is comparable to the artist-designed ambigrams.

\begin{figure}[t]
\centering
\includegraphics[width=0.98\linewidth]{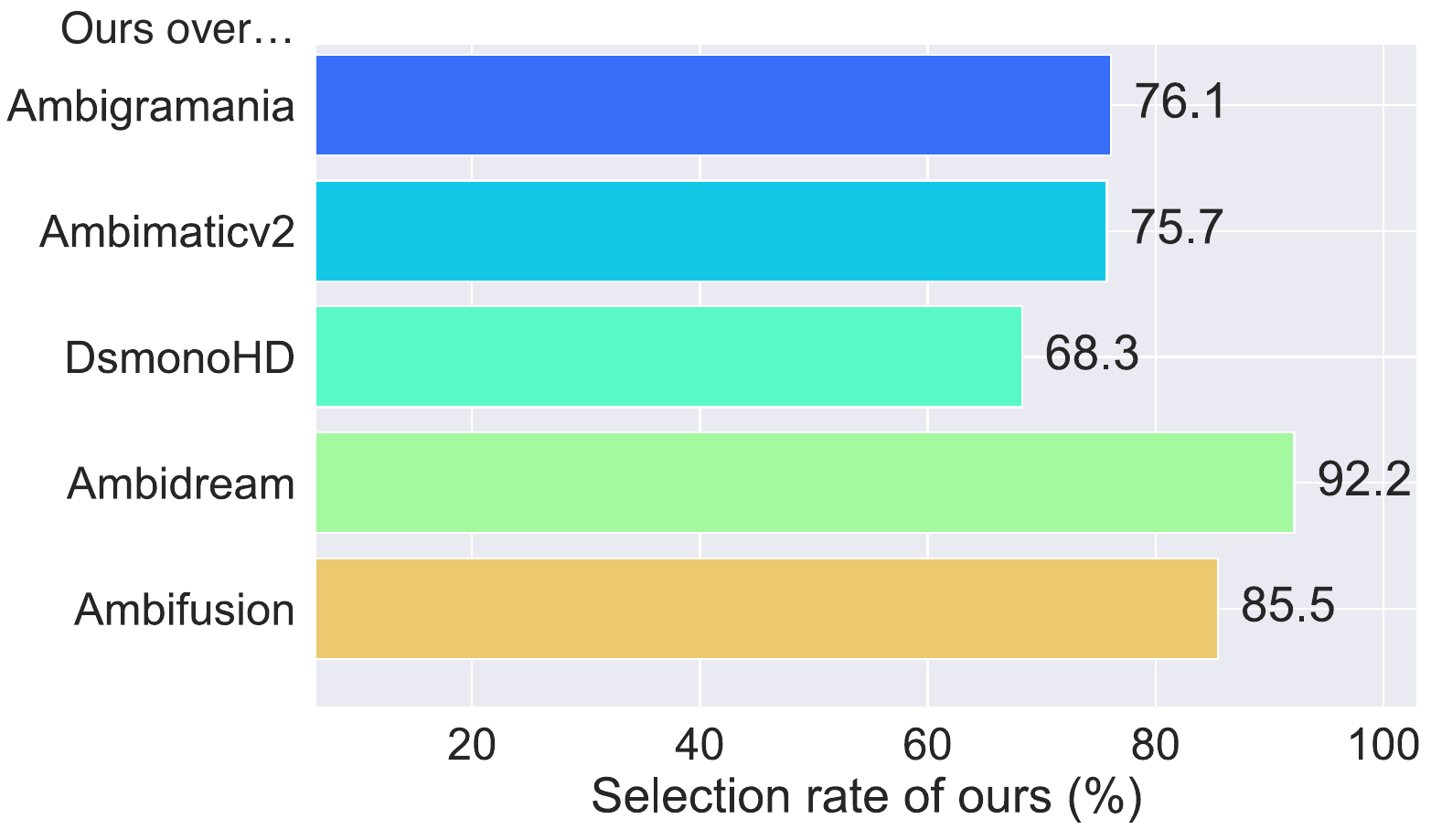}
\vspace{-0.25cm}
\caption{User study results, where we ask the participant to select between generations from our method vs. baselines'. A higher selection rate indicates that our method is more favorable.
}
\vspace{-0.25cm}
\label{fig:user_study}
\end{figure}
\subsection{Qualitative comparisons}
In~\figref{fig:comparison} and~\ref{fig:comparison2}, we provide ambigram designs drawn by artists and generated by AI-based methods. We observe that generations from our approach are readable and aesthetically pleasing, comparable to the artist-designed fonts. For symmetric letters,~\eg, the `H' in ``the'' of~\figref{fig:comparison} our approach maintains the symmetry as done by the artist, while Ambidream and Ambifusion do not. 

Next, we observe that our approach generates designs that are similar in spirit to the ones designed by artists, \eg, the `B' $\leftrightarrow$ `E' of the ``Base'' ambigram in~\figref{fig:comparison}, which leaves the left side of the ``B'' open such that it can be viewed as an `E' when rotated. Our method also has the flexibility of automatically choosing between upper/lower cases, \eg, the middle two `R's in "Correct" of~\figref{fig:comparison2} which made the overall ambigram more legible. 

Finally, we observe that Ambidream generates ambigrams that are difficult to read, yet, it achieves a competitive accuracy in~\tabref{tab:OCR_metric}. We suspect that because Ambidream is distilling from a letter classifier, which biases it towards higher accuracy but does not correlate with human perception. This motivates us to conduct a user study to compare the quality of the generations instead relying solely on TrOCR.

\myparagraph{User study results.}
To quantify the qualitative comparison, we conducted a user study by asking the participants to select between pairs of generations that they find more aesthetically pleasing and legible to the reference word. Specifically, we show them the referenced word, a generation from our method (Font 1), and a generation from one of the baselines. All the choices and order of the words are shuffled for each user. For each ambigram font, we show 20 comparisons. In total, 30 people participated in the survey
and the result is summarized in~\figref{fig:user_study}. We observe that our approach is more favorable when compared to the baselines, with Dsmono being the most competitive against ours and Ambidream being the least. %
Overall, the ranking of the baselines is consistent with the quantitative results from TrOCR, except for Ambidream which distills from a letter classifier.

\begin{figure}[t]
    \centering
    \setlength{\tabcolsep}{1.2pt}
    \renewcommand{\arraystretch}{1.8}
    \begin{tabular}{cl}
    `A' $\leftrightarrow$ `G' &
    \begin{tabular}{*{6}{c}}
        \includegraphics[width=0.05\textwidth]{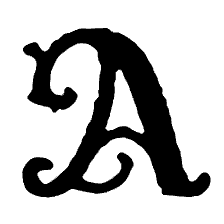} &
        \includegraphics[width=0.05\textwidth]{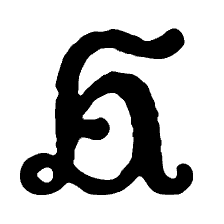} &
        \includegraphics[width=0.05\textwidth]{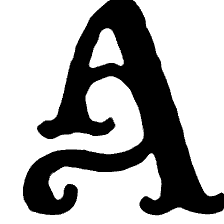} &
        \includegraphics[width=0.05\textwidth]{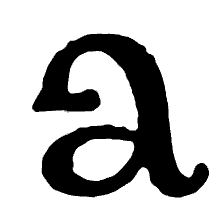} &
        \includegraphics[width=0.05\textwidth]{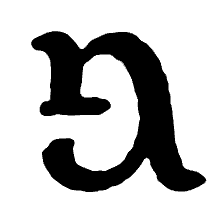} &
        \includegraphics[width=0.05\textwidth]{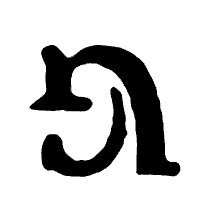} \\
    \end{tabular} \\
    `A' $\leftrightarrow$ `Q' &
    \begin{tabular}{*{6}{c}}
       \includegraphics[width=0.05\textwidth]{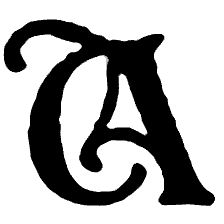} &
        \includegraphics[width=0.05\textwidth]{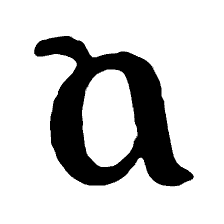} &
        \includegraphics[width=0.05\textwidth]{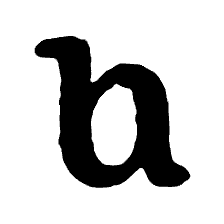} &
        \includegraphics[width=0.05\textwidth]{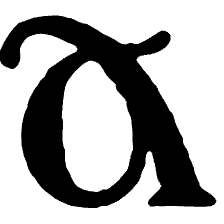} &
        \includegraphics[width=0.05\textwidth]{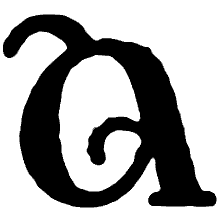} &
        \includegraphics[width=0.05\textwidth]{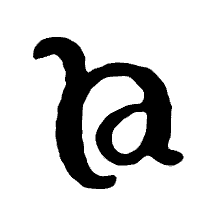} \\
    \end{tabular} \\
    `B' $\leftrightarrow$ `Q' &
    \begin{tabular}{*{6}{c}}
       \includegraphics[width=0.05\textwidth]{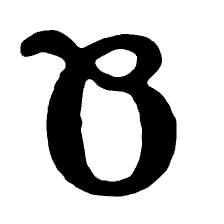} &
        \includegraphics[width=0.05\textwidth]{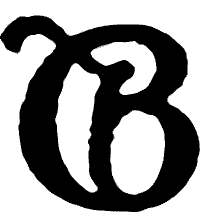} &
        \includegraphics[width=0.05\textwidth]{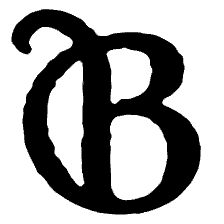} &
        \includegraphics[width=0.05\textwidth]{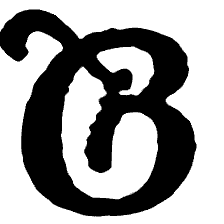} &
        \includegraphics[width=0.05\textwidth]{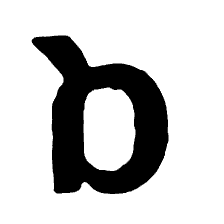} &
        \includegraphics[width=0.05\textwidth]{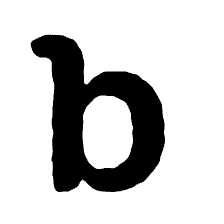} \\
    \end{tabular} \\
    \end{tabular}
    \vspace{-0.4cm}
    \caption{Illustration of diverse generations. Our method generates diverse fonts for a given pair of inputs. Notice that the fonts contain a mix of lower or upper case letters that is determined automatically depending on the legibility.
    }
    \vspace{-0.2cm}
    \label{fig:diversity}
\end{figure}

\myparagraph{Diversity of our approach.}
Our approach is capable of generating diverse designs. The different design arises from changing the initial font, the reference font style that guides the generation process, the alignment strategy, and the randomness from the diffusion model. In~\figref{fig:diversity}, we show examples illustrating the multiple designs for ambigrams between single letters. We observe variations across the style of the generated font, as well as, the choice of using lower or upper case letters.

\subsection{Ablation studies}
We conducted several ablation studies validating the importance of the choices made in our proposed method.

\begin{figure}[t]
\centering
\begin{tabular}{ccc}
`a' $\leftrightarrow$ `a' & `e' $\leftrightarrow$ `q' & `s' $\leftrightarrow$ `e'\\
\includegraphics[width=0.28\linewidth]{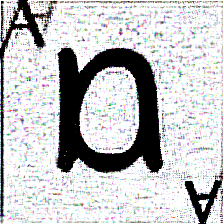} & 
\includegraphics[width=0.28\linewidth]{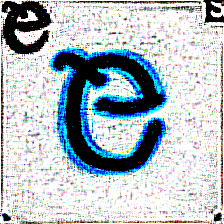} &
\includegraphics[width=0.28\linewidth]{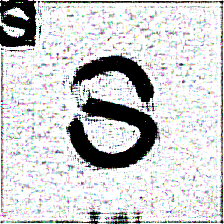}\\
\end{tabular}
\vspace{-0.25cm}
\caption{Directly optimizing pixel intensities. As pixel space has more degree of freedom, we observe that extra letters are placed at the corners of the image. Additionally, the background is noisy, and thus the generations are not readily usable as fonts.
}
\vspace{-0.2cm}
\label{fig:svg_vs_pixel}
\end{figure}

\myparagraph{Ablating vector representation.}
To ablate the choice of using a vector representation, we instead directly optimize the pixel intensities using our method. In~\figref{fig:svg_vs_pixel}, we show the generated results. We observe that the generation leads to multiple duplicated letters that are not centered and have noisy backgrounds. We believe that using a vector representation reduces the \textit{space} %
of what can be generated, which in turn improves the quality.

\begin{figure}[t]
    \centering
    \setlength{\tabcolsep}{3pt}
    \renewcommand{\arraystretch}{1.}
    \begin{tabular}{ccccc}
    \specialrule{.15em}{.05em}{.05em}
    & ``LETTER'' $\leftrightarrow$ ``LETTER'' 
    & ``ROOM'' $\leftrightarrow$ ``ROOM''  \\
    
    \hline
    (a)
    & \raisebox{-.45\height}{\includegraphics[height=0.62cm]{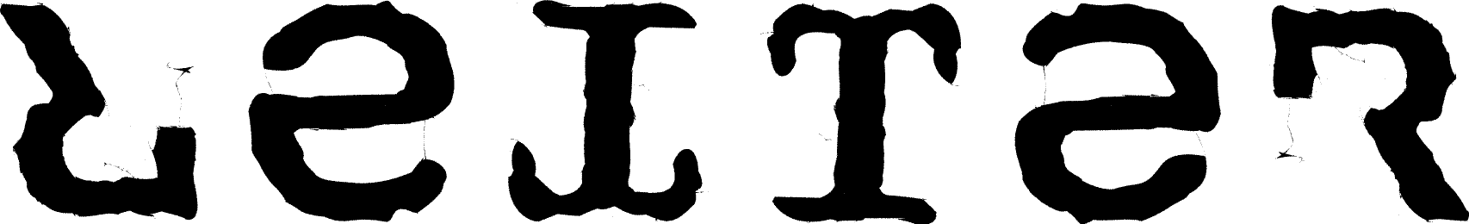}}
    & \raisebox{-.35\height}{\includegraphics[height=0.62cm]{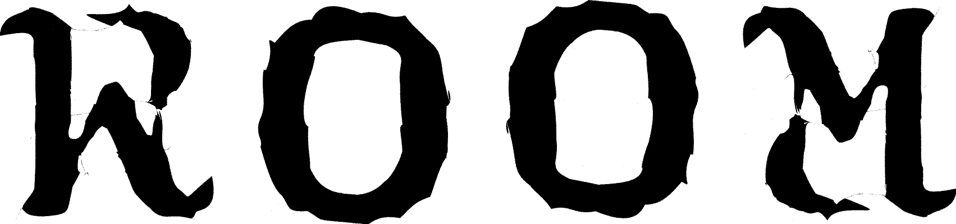}}
    \vspace*{.5mm}
    \\
    (b)
    & \raisebox{-.35\height}{\includegraphics[height=0.62cm]{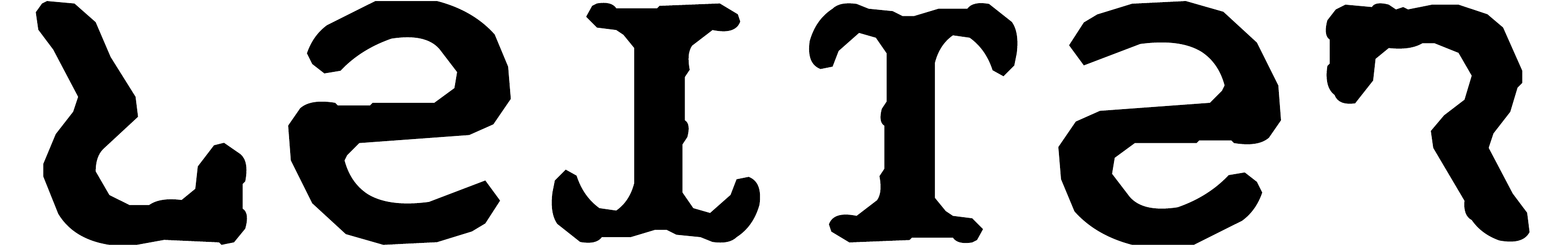}}
    & \raisebox{-.35\height}{\includegraphics[height=0.62cm]{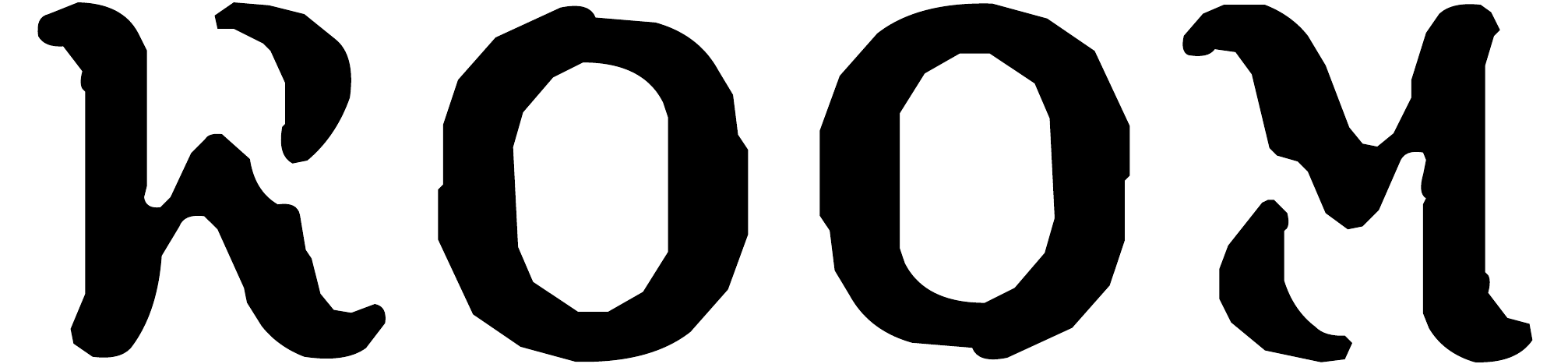}}
    \vspace*{.5mm}\\
    (c)
    & \raisebox{-.45\height}{\includegraphics[height=0.62cm]{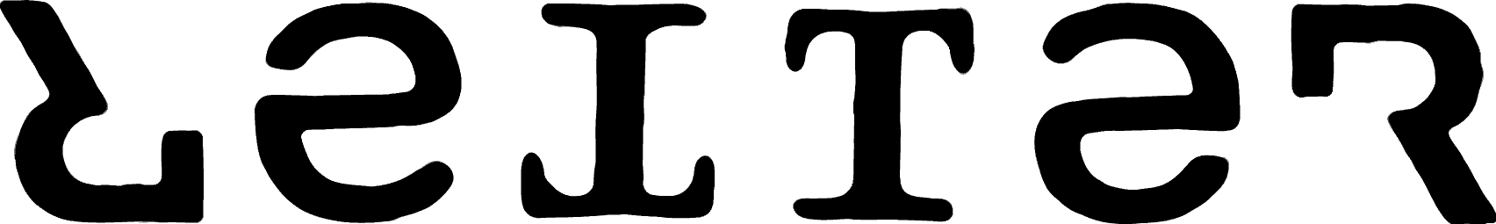}}
    & \raisebox{-.35\height}{\includegraphics[height=0.62cm]{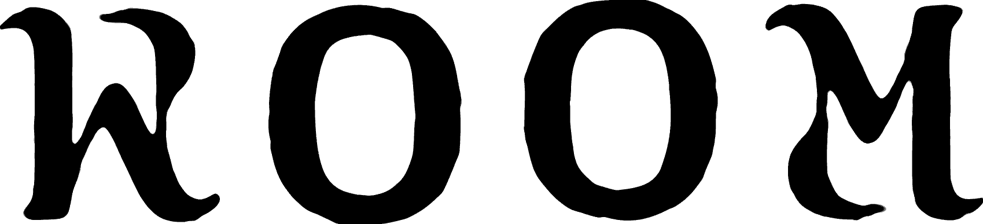}}
    \vspace{.5mm}\\
    \specialrule{.15em}{.05em}{.05em}
    \end{tabular}
    \vspace{-0.2cm}
    \caption{Illustration of generations at different stages of the method.
    \textbf{(a)} Only letter-level optimization
    \textbf{(b)} after post-processing, \textbf{(c)} after word-level optimization. The quality of the ambigrams improves after each step.
    }
    \vspace{-0.3cm}
    \label{fig:joint_opt_compare}
\end{figure}

\myparagraph{Ablating word-level optimization.}
Our method performs letter-level optimization ($\gL_{\tt Letr}$ in \equref{eq:loss_letter}), 
followed by an word-level optimization ($\gL_{\tt Word}$ in \equref{eq:loss_word}). 
In~\figref{fig:joint_opt_compare}, we show the result after letter-level optimization, after post-processing, and the final result. As can be seen, the results in row (c) have the highest quality, validating the effectiveness of the word-level optimization.

\begin{figure}[t]
    \centering
    \setlength{\tabcolsep}{3pt}
    \begin{tabular}{ccccc}
    \specialrule{.15em}{.05em}{.05em}
    & None & Typewriter & Quarterly & Genzsch\\
    \hline
    `a' $\leftrightarrow$ `E'
    & \raisebox{-.35\height}{\includegraphics[height=1.25cm]{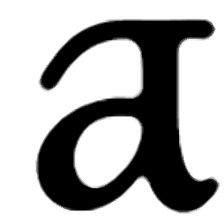}}
    & \raisebox{-.35\height}{\includegraphics[height=1.3cm]{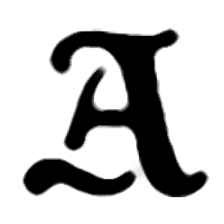}}
    & \raisebox{-.35\height}{\includegraphics[height=1.3cm]{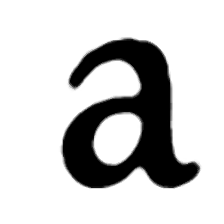}}
    & \raisebox{-.35\height}{\includegraphics[height=1.3cm]{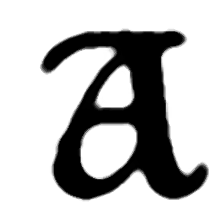}}
    \\
    `C' $\leftrightarrow$ `E'
    & \raisebox{-.35\height}{\includegraphics[height=1.25cm]{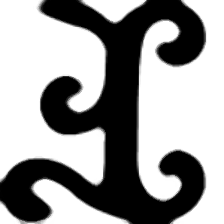}}
    & \raisebox{-.35\height}{\includegraphics[height=1.3cm]{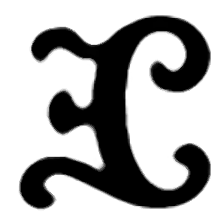}}
    & \raisebox{-.35\height}{\includegraphics[height=1.3cm]{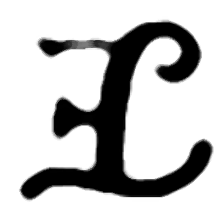}}
    & \raisebox{-.35\height}{\includegraphics[height=1.3cm]{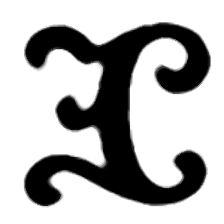}}
    \vspace{1mm}\\
    \specialrule{.15em}{.05em}{.05em}
    \end{tabular}
    \vspace{-0.2cm}
    \caption{{Ablation of style loss with different reference font.}
    }
    \vspace{-0.3cm}
    \label{fig:font_control}
\end{figure}

\myparagraph{Ablating style loss $\gL_{\tt Style}$.}
In~\figref{fig:font_control}, we provide generations with different reference fonts using the style loss. As can be observed, different input reference fonts lead to different styles across the same ambigram. 

\begin{figure}[t]
    \centering
    \setlength{\tabcolsep}{1pt}
    \begin{tabular}{*{12}{cc}}
    \specialrule{.15em}{.05em}{.05em}
    $\lambda$ & 1.0 & 0.9 & 0.8 & 0.7 & 0.6 & 0.5 & 0.4 & 0.3 & 0.2 & 0.1 & 0.0\\ 
    \midrule
    &  \includegraphics[width=0.035\textwidth]{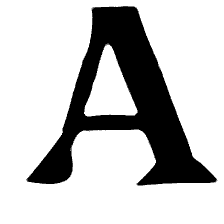} &
        \includegraphics[width=0.035\textwidth]{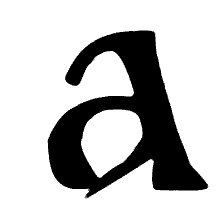} &
        \includegraphics[width=0.035\textwidth]{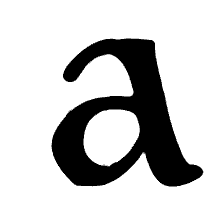} &
        \includegraphics[width=0.035\textwidth]{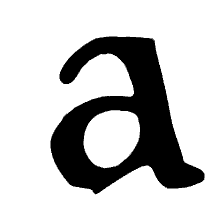} &
        \includegraphics[width=0.035\textwidth]{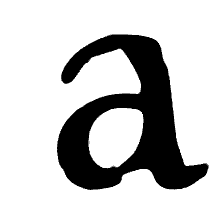} &
        \includegraphics[width=0.035\textwidth]{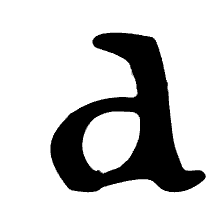} &
        \includegraphics[width=0.035\textwidth]{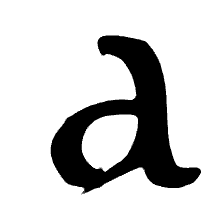} &
        \includegraphics[width=0.035\textwidth]{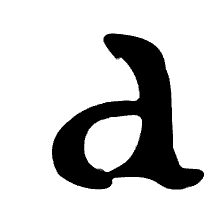} &
        \includegraphics[width=0.035\textwidth]{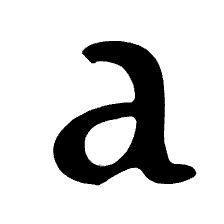} &
        \includegraphics[width=0.035\textwidth]{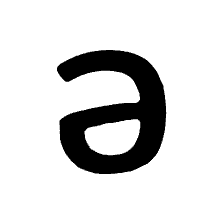} &
        \includegraphics[width=0.035\textwidth]{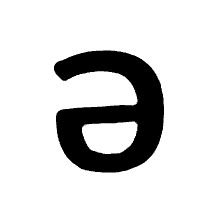} \\
    &
        \includegraphics[width=0.035\textwidth]{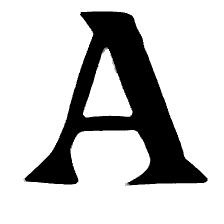} &
        \includegraphics[width=0.035\textwidth]{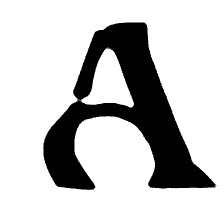} &
        \includegraphics[width=0.035\textwidth]{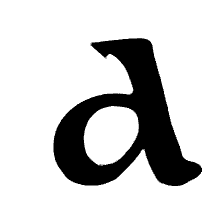} &
        \includegraphics[width=0.035\textwidth]{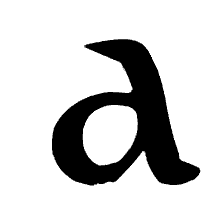} &
        \includegraphics[width=0.035\textwidth]{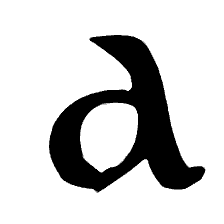} &
        \includegraphics[width=0.035\textwidth]{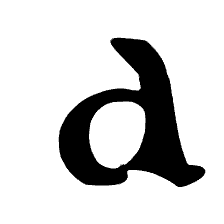} &
        \includegraphics[width=0.035\textwidth]{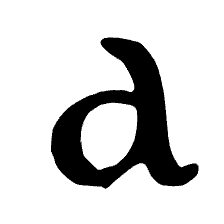} &
        \includegraphics[width=0.035\textwidth]{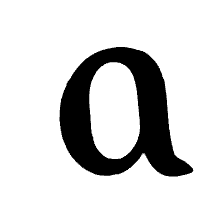} &
        \includegraphics[width=0.035\textwidth]{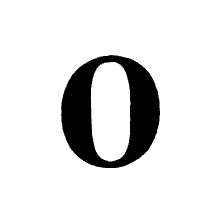} &
        \includegraphics[width=0.035\textwidth]{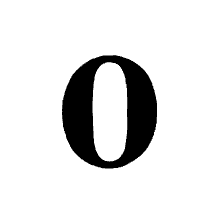} &
        \includegraphics[width=0.035\textwidth]{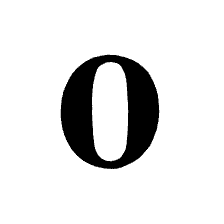} \\
    &
        \includegraphics[width=0.035\textwidth]{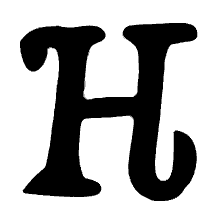} &
        \includegraphics[width=0.035\textwidth]{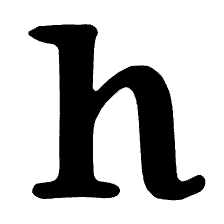} &
        \includegraphics[width=0.035\textwidth]{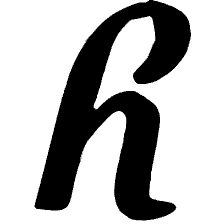} &
        \includegraphics[width=0.035\textwidth]{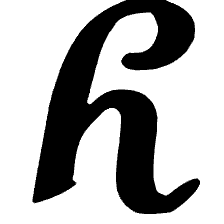} &
        \includegraphics[width=0.035\textwidth]{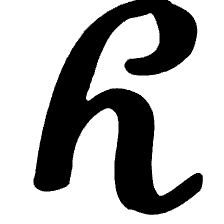} &
        \includegraphics[width=0.035\textwidth]{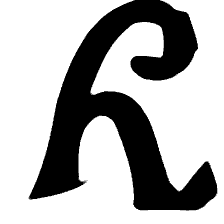} &
        \includegraphics[width=0.035\textwidth]{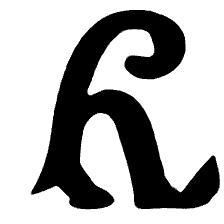} &
        \includegraphics[width=0.035\textwidth]{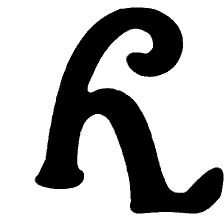} &
        \includegraphics[width=0.035\textwidth]{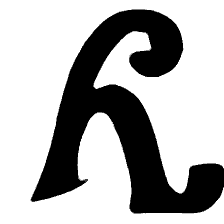} &
        \includegraphics[width=0.035\textwidth]{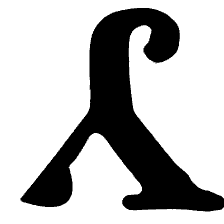} &
        \includegraphics[width=0.035\textwidth]{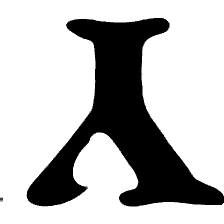} \\
        \specialrule{.15em}{.05em}{.05em}
    \end{tabular}
    \vspace{-0.2cm}
    \caption{Sweep across $\lambda_{\tt Letter}$ in~\equref{eq:loss_letter} which balances the legibility of each letter in the upright and rotated orientation.
    }
    \vspace{-0.4cm}
\label{fig:bias_weight_sweep}
\end{figure}

\myparagraph{Importance of selecting loss weights $\lambda_{\tt Letr}$.}
The loss $\gL_{\tt Letr}$ in~\equref{eq:loss_letter} based DeepFloyd IF has an imbalance between the letters, \ie, it favors certain letters over others. This causes difficulties when choosing $\lambda_{\tt Letr}$. In~\figref{fig:bias_weight_sweep}, we show the generation at different $\gL_{\tt Letr}$. Consider the first row between `A' $\leftrightarrow$ `e', the two viewing orientations are both legible at $\lambda_{\tt Letr}=0.2$. On the other hand, the sweet spot for `H' $\leftrightarrow$ `Y' is with $\lambda_{\tt Letr}=0.6$. This emphasizes the need to automatically select $\lambda_{\tt Letr}$ as proposed in our method.

\begin{figure}
    \centering
    \begin{tabular}{*{6}{c}}
    \specialrule{.15em}{.05em}{.05em}
    {\tt Ver.} & `a' $\leftrightarrow$ `a' & `a' $\leftrightarrow$ `b' & `a' $\leftrightarrow$ `c' & `a' $\leftrightarrow$ `d'
    \\
    \hline
    \raisebox{10pt}{\tt v1-5} &
    \includegraphics[width=0.06\textwidth,trim={2cm 1.5cm 1cm 1.5cm},clip]{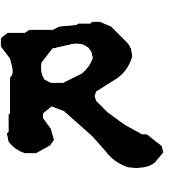} &
    \includegraphics[width=0.06\textwidth,trim={2cm 1.5cm 1cm 1.5cm},clip]{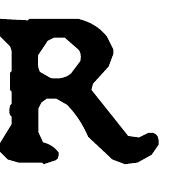} &
    \includegraphics[width=0.06\textwidth,trim={2cm 1.5cm 1cm 1.5cm},clip]{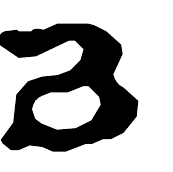} &
    \includegraphics[width=0.06\textwidth,trim={2cm 1.5cm 1cm 1.5cm},clip]{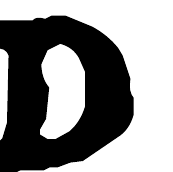} \\
    \raisebox{10pt}{\tt v2-1} &
    \includegraphics[width=0.06\textwidth]{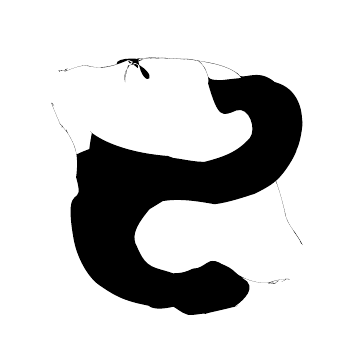} &
    \includegraphics[width=0.06\textwidth]{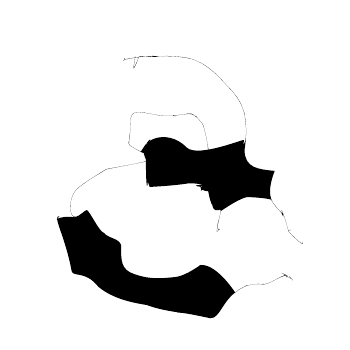} &
    \includegraphics[width=0.06\textwidth]{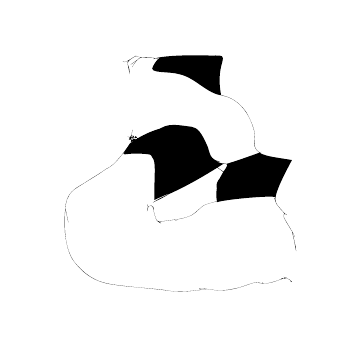} &
    \includegraphics[width=0.06\textwidth]{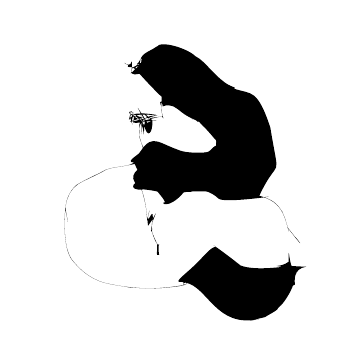} \\
    \specialrule{.15em}{.05em}{.05em}
    \end{tabular}
    \vspace{-0.2cm}
    \caption{{Results when distilling with StableDiffusion.} Observe that both v1-5 and v2-1 version of StableDiffusion is unable to produce a coherent letter. %
    }
    \vspace{-0.2cm}
    \label{fig:stable_diffusion_vs_IF}
\end{figure}

\myparagraph{Choice of pre-trained diffusion model.}
We have also experimented with using other pre-trained diffusion models other than Deep Floyd IF. We provide results from StableDiffusion~\cite{rombach2022high} v1-5 and v2-1 in~\figref{fig:stable_diffusion_vs_IF}. %
Overall, we found that StableDiffusion models struggle to understand/generate ambigrams.
This is consistent with the observation that StableDiffusion has difficulties generating text when prompted. On the other hand, DeepFloyde IF uses a different text encoder which is found to improve text understanding~\cite{deep-floyd-if}.

\subsection{Proof of concept for unequal length ambigrams}
\begin{figure}[t]%
    \centering
    \renewcommand{\arraystretch}{0.1}
    \begin{tabular}{cc}
    \multicolumn{2}{c}{``Stir'' $\leftrightarrow$ ``Sup''}\\
    \includegraphics[height=1.5cm]{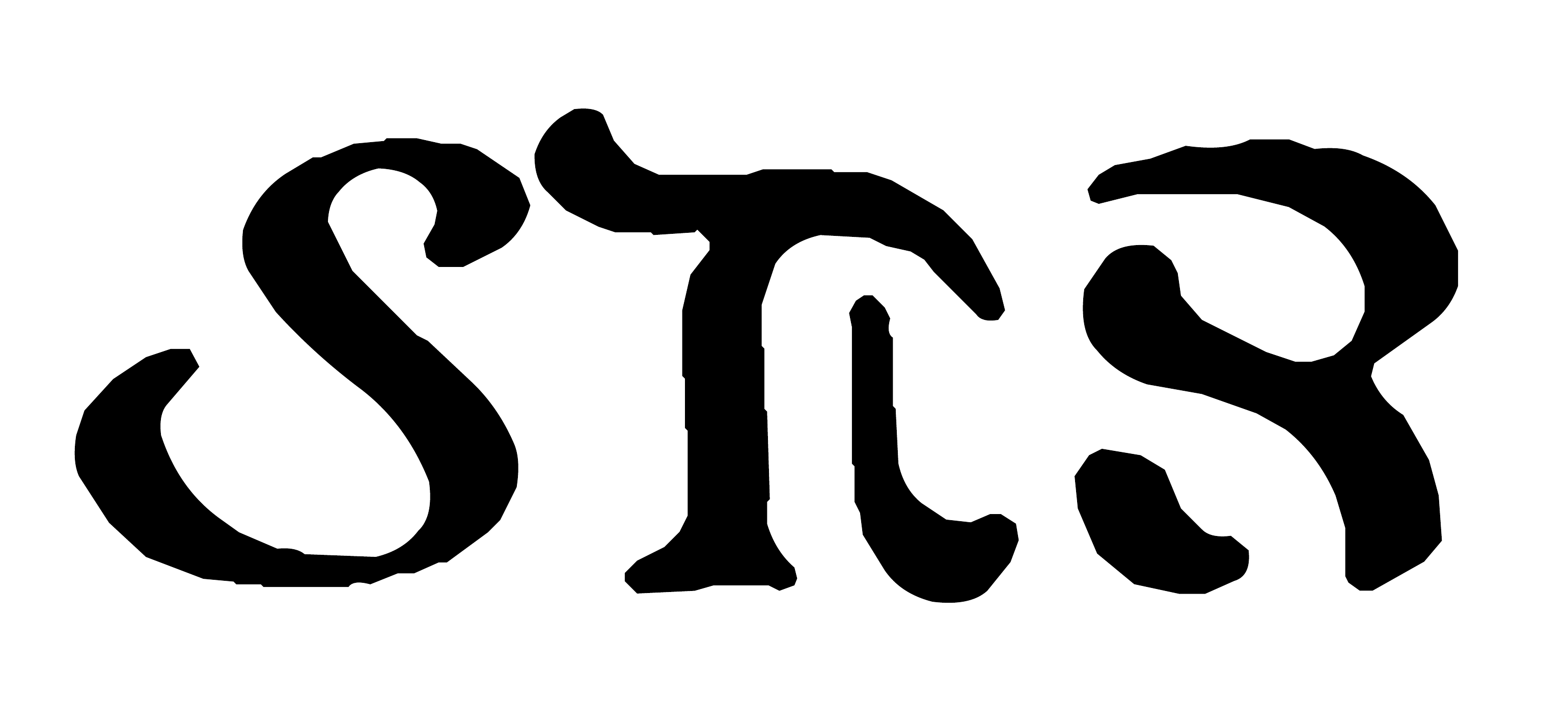} & 
    \includegraphics[height=1.5cm, angle=180,origin=c]{imgs/two_to_one_examples/stir_to_sup}\\
    \multicolumn{2}{c}{``Fight'' $\leftrightarrow$ ``Easy''}\\
     \includegraphics[height=1.5cm]{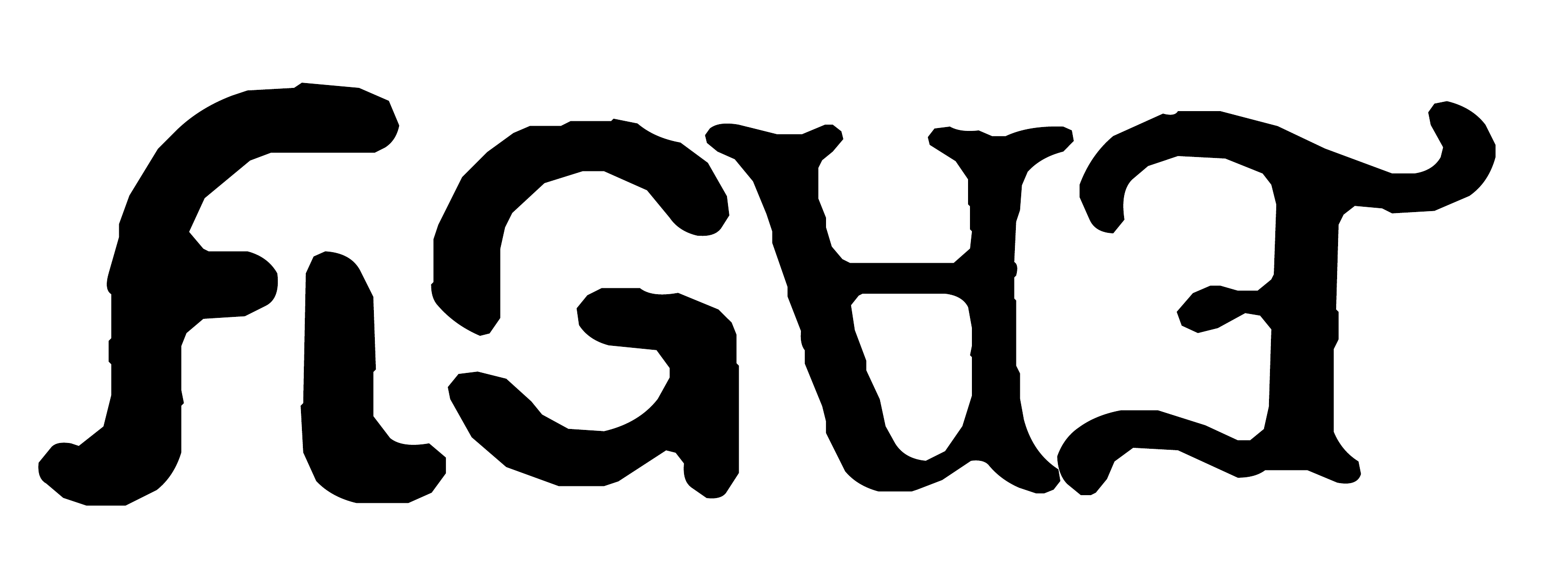}  & 
     \includegraphics[height=1.5cm, angle=180,origin=c]{imgs/two_to_one_examples/fight_to_easy}\\
    \end{tabular}
    \vspace{-0.25cm}
    \caption{Proof of concept for unequal length ambigrams, \ie, the number of letters in one orientation differs from the other. This setting is not possible from existing methods due to the one-to-one mapping assumption during their design process.
    }
    \vspace{-0.2cm}
    \label{fig:unbalance_word_length}
\end{figure}

Thus far in the paper, we have shown ambigrams $\va \leftrightarrow \vb$ where the words $\va$ and $\vb$ have the same length. This is because existing methods use a letter-to-letter design process and therefore are unable to support unequal lengths.
We now showcase that our method can generalize to designing ambigrams of \textit{unequal lengths} by encouraging, via $\gL_{\tt Letr}$ in~\equref{eq:loss_letter}, a glyph to be read as one letter and two letters in the other orientation through modifying the text prompt.
In~\figref{fig:unbalance_word_length}, we show examples of ambigrams ``Stir'' $\leftrightarrow$ ``Sup'', and ``Fight'' $\leftrightarrow$ ``Easy''. We note that these results are a proof of concept. The optimization currently involves manual tuning on the alignment and balancing of the loss terms for each ambigram. We aim to tackle this very challenging task of unequal length generation in future work. 


\section{Conclusion}
In this paper, we propose a novel method for ambigram generation by distilling DeepFloyd IF, a pre-trained diffusion model. Unlike existing methods that only consider designs at the letter level, our approach considers the quality of the ambigram at both the letter and word levels. To our knowledge, we are also the first to consider word-level evaluation of ambigrams. Experimental results show our method's superior performance, achieving more than an 11.2\% increase in accuracy and at least a 41.9\% reduction in edit distance over the baselines. These quantitative results are also supported by a user study. %

\clearpage
{
\small
\bibliographystyle{ieeenat_fullname}
\bibliography{ref}
}

\end{document}